\documentclass{article}

\usepackage{PRIMEarxiv}

\usepackage[utf8]{inputenc} 
\usepackage[T1]{fontenc}    
\usepackage{hyperref}       
\usepackage{url}            
\usepackage{booktabs}       
\usepackage{amsfonts}       
\usepackage{nicefrac}       
\usepackage{microtype}      
\usepackage{lipsum}
\usepackage{fancyhdr}       
\usepackage{graphicx}       
\graphicspath{{media/}}     

\usepackage{lineno}
\usepackage{amsmath}
\usepackage{amssymb}
\usepackage{mathtools}
\usepackage{amsthm}
\usepackage{algorithm} 
\usepackage{algorithmic} 
\floatname{algorithm}{Algorithm}

\def\mathbi#1{\textbf{\em #1}}
\usepackage{xcolor}
\usepackage{subfigure}
\usepackage{caption}
\usepackage{graphicx,xcolor} 
\usepackage[framemethod=tikz]{mdframed}
\usepackage{xcolor}
\def\mathbi#1{\textbf{\em #1}}
\usepackage{caption}
\usepackage{sidecap}
\usepackage{breqn}
\usepackage{bm}

\newcommand{\DesignVar}{\mathbi{x}}

\newcommand{\DesignSpace}{\mathcal{X}}
\newcommand{\ObjFun}{f}

\newcommand{\MinDesignVar}{x^{*}}
\newcommand{\LevFid}{l}
\newcommand{\MaxLevFid}{L}
\newcommand{\Dim}{d}
\newcommand{\IterOpt}{i}
\newcommand{\Budget}{B}
\newcommand{\Expectation}{\mathbb{E}}

\newcommand{\NoisyObserv}{y}
\newcommand{\IndNumObs}{n} 
\newcommand{\Dataset}{\mathcal{D}}
\newcommand{\NumObs}{N} 

\newcommand{\StandDevNoise}{\sigma_{\epsilon}}
\newcommand{\NorDist}{\mathcal{N}}
\newcommand{\MeanFunGP}{\mu}
\newcommand{\CovFunGP}{\kappa}
\newcommand{\StandDevGP}{\sigma}
\newcommand{\KernelMatrix}{\mathbi{K}} 
\newcommand{\IdMatrix}{\mathbi{I}}

\newcommand{\ConsFact}{\varrho}
\newcommand{\Discrep}{\delta}
\newcommand{\AF}{U}
\newcommand{\EI}{EI}

\newcommand{\CompCost}{\lambda}
\newcommand{\CumDistFun}{\Phi}

\newcommand{\Probability}{P}
\newcommand{\PhysicsVec}{\boldsymbol{\psi}}
\newcommand{\Improv}{I}

\newcommand{\Mach}{M}
\newcommand{\Weight}{w}

\newcommand{\Residuals}{\mathbi{R}}
\newcommand{\ConsVar}{\mathbi{U}}
\newcommand{\Time}{t}
\newcommand{\Density}{\rho}
\newcommand{\Velocity}{v}

\newcommand{\Energy}{E}
\newcommand{\Source}{\mathbi{Q}}
\newcommand{\ConvFluxes}{\mathbi{F}^c}
\newcommand{\ViscFluxes}{\mathbi{F}^v}
\newcommand{\CompDomain}{\Omega}
\newcommand{\Pressure}{p}
\newcommand{\Temperature}{T}
\newcommand{\Conductivity}{k}

\newcommand{\ViscStress}{\overline{\bm{\tau}}}
\newcommand{\Altitude}{h}
\newcommand{\DragCoef}{C_d}
\newcommand{\LiftCoef}{C_l}
\newcommand{\PitchCoef}{C_m}
\newcommand{\Thick}{t}
\newcommand{\Chord}{c}
\newcommand{\Radius}{r}
\newcommand{\TrailAngle}{\tau}
\newcommand{\Interval}{I}

\newcommand{\discrep}{\gamma} 
\newcommand{\Strain}{S}

\newcommand{\FaultParam}{q}

\newcommand{\Error}{e}

\title{Physics-Aware Multifidelity Bayesian Optimization: a Generalized Formulation}

\author{
 Francesco Di Fiore \\
  Politecnico di Torino\\
  \texttt{francesco.difiore@polito.it} \\
   \And
 Laura Mainini \\
  Imperial College London\\
  Politecnico di Torino\\
  Massachusetts Institute of Technology\\
  \texttt{l.mainini@imperial.ac.uk} \\
}

\begin{document}
\maketitle

\begin{abstract}
The adoption of high-fidelity models for many-query optimization problems is majorly limited by the significant computational cost required for their evaluation at every query. Multifidelity Bayesian methods (MFBO) allow to include costly high-fidelity responses for a sub-selection of queries only, and use fast lower-fidelity models to accelerate the optimization process. State-of-the-art methods rely on a purely data-driven search and do not include explicit information about the physical context. This paper acknowledges that prior knowledge about the physical domains of engineering problems can be leveraged to accelerate these data-driven searches, and proposes a generalized formulation for MFBO to embed a form of domain awareness during the optimization procedure. In particular, we formalize a bias as a multifidelity acquisition function that captures the physical structure of the domain. This permits to partially alleviate the data-driven search from learning the domain properties on-the-fly, and sensitively enhances the management of multiple sources of information. The method allows to efficiently include high-fidelity simulations to guide the optimization search while containing the overall computational expense. Our physics-aware multifidelity Bayesian optimization is presented and illustrated for two classes of optimization problems frequently met in science and engineering, namely design optimization and health monitoring problems.
\end{abstract}

\let\thefootnote\relax\footnotetext{Full paper: Di Fiore, Francesco, and Laura Mainini. "Physics-aware multifidelity Bayesian optimization: A generalized formulation." Computers \& Structures 296 (2024): 107302. https://doi.org/10.1016/j.compstruc.2024.107302}



\section{Introduction}
\label{s: Intro}

Optimization problems are ubiquitous in science and engineering applications \cite{martins2021engineering}. Those also include the support to engineering tasks that are in increasing demand to meet sustainability goals such as the identification of the best design configurations to maximize the performance and minimize the environmental impact of novel engineering solutions, and the detection and identification of damages or faults to monitor the health condition of complex systems to maximize their useful life and minimize waste of resources. 

Over the last decades, the increase of computing power and the advances in computational modelling capabilities made available computer-based models for the accurate analysis and simulation of complex physical systems. This is the case of computational schemes for the numerical solution of governing partial differential equations as computational fluid dynamic solvers to represent viscous fluids, and finite element methods for the analysis of mechanical structures, heath transfer and electromagnetic phenomena. In principle, this computer-based representations can provide a remarkable contribution to enhance the search and identification task in simulation-based optimization. Unfortunately, the extensive adoption of these high-fidelity models during the optimization procedure is hampered by the significant computational cost and time required for their evaluation, potentially in the order of months for a single evaluation on high performance computing platforms. This issue becomes more challenging for many-query optimization problems where the demand for model evaluations grows exponentially with the number of parameters to optimize. 

The use of low-fidelity models constitutes a popular approach to reduce the computational resources associated with the solution of optimization problems. Low-fidelity representations introduce assumptions about the physics and/or approximate the solution of the governing equations, and relief the computational expenditure for the evaluation of the response of the system. On one hand, the use of low-fidelity models allows to efficiently acquire and exploit a massive amount of information; on the other hand, their adoption results in a reduction of the overall time required for optimization. However, the main drawback of this strategy is enclosed in the simplified modeling approach which might not be adequate to capture complex physical phenomena that characterize many advanced technological applications, providing inaccurate responses that potentially lead to the suboptimal identification of solutions.

Multifidelity methods offer the opportunity to efficiently include high-fidelity models during the optimization procedure, combining information elicited from a library of models hierarchically ordered accordingly to the accuracy of the response and associated expense of the computations \cite{forrester2007multi, park2017remarks, peherstorfer2018survey, beran2020comparison}. These methodologies use fast low-fidelity models to speedup the search procedure, and refine the solution identified through of a principled selection of high-fidelity information. The goal is to accelerate the identification of optimal solutions while managing a trade-off between accuracy and computational cost. Multifidelity frameworks have been applied to solve a variety of optimization problems. Examples include and are not limited to the design optimization of aircraft \cite{thelen2022multi}, vessels \cite{tezzele2023multifidelity}, and hybrid vehicle \cite{anselma2020multidisciplinary}, and health assessment of civil structures \cite{lai2022building}, composite wings \cite{makkar2022machine} and industrial systems \cite{lai2022building}.

Multifidelity Bayesian Optimization (MFBO) represents a class of computational techniques that relies on a Bayesian framework to address optimization problems, and combines data from sources of information at different levels of fidelity \cite{viana2014special, guo2018analysis, song2019general}. The Bayesian methodology defines a scheme for the optimization of black-box functions where a probabilistic surrogate model is updated at each iteration using evaluations of models prescribed by an acquisition function \cite{movckus1975bayesian, snoek2012practical, frazier2018bayesian}. The surrogate model provides an approximation of the objective function over the domain while the acquisition function determines a policy to measure the utility of evaluating the objective function in certain locations of the domain. Whether a library of models is available, multifidelity Bayesian optimization synthesizes data elicited through multiple sources of information into a unique surrogate model, and guides the search through an adaptive sampling scheme based on a multifidelity acquisition function that selects the most promising location and the associated level of fidelity to query. 
 
MFBO has been widely adopted to solve optimization problems in science and engineering. Meliani et al. \cite{meliani2019multi} propose a multifidelity Bayesian technique for high-dimensional design optimization problems, and demonstrate the methodology for the aerodynamic shape optimization of an airfoil outperforming a standard single-fidelity Bayesian algorithm. Tran et al. \cite{tran2020multi} develop a multifidelity Bayesian framework for the inference of the optimal chemical composition in material science, dedicating particular attention to the identification of the optimal concentration of components achieving the desired bulk modulus for ternary random alloys. Serani et al. \cite{serani2019adaptive} propose a multifidelity approach based on stochastic radial basis functions to include expensive computational fluid dynamic simulations in global design optimization, and demonstrate the methodology for the optimization of a NACA hydrofoil and a destroyer-type vessel for naval engineering applications. Perdikaris and Karniadakis \cite{perdikaris2016model} adopt a multifidelity Bayesian strategy to solve inverse problems in haemodynamics through the identification of system parameters that characterize physiologically correct blood flow simulations.

All these multifidelity frameworks rely on fully data-driven adaptive samplings informed exclusively through probabilistic data extracted from the surrogate model. Accordingly, the search procedure has to learn entirely from data both the surrogate model and the characterization of the discrepancies -- frequently non-linear -- between the different models over the entire physical domain. This can still require a large amount of high-fidelity information to capture the modeling correlations for each level of fidelity available, and results in intensive computations associated with the massive evaluations of accurate numerical models. In the engineering context, prior knowledge about these discrepancies is at disposal, either because it is formalized by the governing equations that represent the physics of the system or because it derives from the know-how of experts about the distinguishing physical phenomena characterizing the system behaviour. In these optimization scenarios, the introduction of the physics-awareness during the search procedure could lead to a principled and efficient use of high-fidelity data according to the prior knowledge about the physics of the system.

This paper aims at capturing this opportunity and proposes an original physics-aware multifidelity Bayesian optimization that incorporates prior domain knowledge to further improve and accelerate the optimization search in multifidelity settings. This permits to partially alleviate the data-driven search from the characterization of the domain structure while trying to identify the optimal combination of parameters that minimizes the objective function. In previous works, we introduced preliminary approaches to capture this prior/expert knowledge in the form of domain-awareness \cite{di2021multifidelity, di2022non}. 

This work proposes a formal generalization of this sort of physics-aware reasoning for a broader relevance and direct applicability to multiple classes of optimization problems in science and engineering. The proposed framework adopts the multifidelity Gaussian process regression to model the belief about the objective function over the entire domain, which is progressively updated through an original acquisition function based on the multifidelity expected improvement. The multifidelity acquisition function is distinctively shaped to combine (i) data-driven information extracted from the surrogate model and (ii) prior/expert knowledge about the structure of the domain encapsulated during the search through a physics-aware utility function. This form of awareness allows to define an adaptive sampling scheme that efficiently manages different sources of information, targeting the balance between computational cost and accuracy demanded along the optimization search. 

We demonstrate our physics-aware multifidelity framework for two popular families of optimization problems, namely the identification of the best combination of design parameters to maximize systems performance, and the non-destructive identification of systems damages or faults. In the first case, we consider the specific example of the aerodynamic design optimization of a transonic airfoil and the prior domain knowledge concerns the sensitivity of the fluid dynamic regime to the variation of the Mach number. The health monitoring problem demands for the assessment of the fault condition of a composite structure, and the prior domain knowledge relates to the effects that can be observed for different extensions of the damage. The two forms of scientific and expert domain knowledge are formulated as physics-aware utility functions that bias the query of numerical models depending on the structure of the domain.

This manuscript is organized as follows: Section \ref{s: probsetup} describes the multifidelity Bayesian scheme and Section \ref{s: Methodology} presents our original physics-aware multifidelity Bayesian framework in detail. In Section \ref{s: AeroOptimization} and Section \ref{s: structprob}, the proposed methodology is applied for an aerodynamic design problem and a structural health monitoring test case, respectively, and the formalization of our physics-aware technique together with the results are discussed. Finally, Section \ref{s:Conclusions} provides concluding remarks.

\section{Multifidelity Bayesian Optimization: Problem Setup}
\label{s: probsetup}

The goal of optimization problems is to identify a combination of parameters $\MinDesignVar$ that minimizes an unknown objective function $\ObjFun(\DesignVar)$, and is mathematically described as:

\begin{equation} \label{e: optproblem}
    \MinDesignVar = \min_{\DesignVar \in \DesignSpace} \ObjFun(\DesignVar)
\end{equation}

\noindent where $\DesignSpace \in \mathbb{R}^{\Dim}$ is the domain of the objective function. Frequently in these settings, the objective function is a black-box input/output relationship whose analytical form and derivatives are not explicitly available.

Bayesian optimization (BO) is a popular surrogate-based method that uses the Bayesian probability theory to address the optimization of black-box objective functions \cite{movckus1975bayesian, snoek2012practical, frazier2018bayesian}. BO realizes a sequential derivative-free optimization procedure based on two core elements: a stochastic surrogate model that approximates the response of numerical models and emulates the distribution of the objective function over the domain, and an acquisition function that determines the optimal sequence of future samples to be evaluated. The overall computational efficiency of BO can be improved combining data from multiple sources of information at different levels of fidelity $\{\ObjFun^{(1)},...,\ObjFun^{(\LevFid)},..., \ObjFun^{(\MaxLevFid)}\}$: usually in this set the higher the level of fidelity, the more accurate and yet expensive its evaluation. In multifidelity Bayesian optimization, both the stochastic emulator and the acquisition function must accommodate the multiple sources available to compute (approximated) observations of the objective function. In particular, the surrogate model approximates the objective function by synthesizing the information from multiple fidelities into a unique emulator. Commonly, the predictive framework for MFBO is based on the extension of Gaussian processes surrogate models to multiple levels of fidelity through an autoregressive scheme \cite{williams1995gaussian, schulz2018tutorial, kennedy2000predicting}. 

The multifidelity acquisition function defines an adaptive sampling scheme that permits to sequentially decide the best location of the domain and associated source of information to query. This decision making procedure is typically driven entirely by the statistical data in output from the surrogate model. The multitude of multifidelity acquisition functions proposed in literature are often developed leveraging existing acquisition functions adopted in single-fidelity Bayesian optimization, such as the Expected Improvement (EI) \cite{jones1998efficient}, Probability of Improvement (PI) \cite{kushner1964new} and Entropy Search (ES) \cite{hennig2012entropy}. Examples of popular multifidelity acquisition functions include the Multifidelity Expected Improvement (MFEI) \cite{huang2006sequential}, Variable-Fidelity Probability of Improvement (VFPI) \cite{ruan2020variable}, Multifidelity Predictive Entropy Search (MFPES) \cite{zhang2017information}, and Multifidelity Max-Value Entropy Search (MFMES) \cite{takeno2020multi}. The primary difference between these formulations is that the decision making process is realized through different approaches to measure the reward of unknown samples. The goal is to achieve the trade-off between the exploration of the domain $\DesignVar$ in locations where the uncertainty associated with the surrogate model is higher, and the exploitation in regions where the prediction of the emulator indicates that the optimum is likely to be located.

\begin{algorithm}[t!] 
\caption{Multifidelity Bayesian optimization}
\label{a:OptimizationAlgorithm}
\begin{algorithmic}[1]

\REQUIRE  Design space $\DesignSpace \in \mathbb{R}^{\Dim}$, multifidelity objective function $\ObjFun^{(\LevFid)}(\DesignVar)$ and the prior distribution $ \Probability(\ObjFun^{(\LevFid)})$

\ENSURE $\MinDesignVar = \min_{\DesignVar \in \DesignSpace} \ObjFun(\DesignVar)$

\STATE $\Dataset_1 \leftarrow \{ \DesignVar_{\IndNumObs},\ObjFun^{(\LevFid_\IndNumObs)},\LevFid_{\IndNumObs} \}_{\IndNumObs=1}^{\NumObs_1}$ collect initial observations

\STATE $\IterOpt \leftarrow 1$

\REPEAT 

       \STATE Update the posterior distribution $\Probability(\ObjFun^{(\LevFid)} | \Dataset_{\IterOpt})$ using $\Dataset_{\IterOpt}$

       \STATE Maximize the acquisition function $\{\DesignVar_{\IterOpt+1},\LevFid_{\IterOpt+1} \} = \max \AF(\DesignVar, \LevFid)$  

       \STATE Evaluate $\ObjFun^{(\LevFid_{\IterOpt+1})}(\DesignVar_{\IterOpt+1})$ 

       \STATE Augment the dataset $\Dataset_{\IterOpt+1} = \Dataset_{\IterOpt} \cup  \{ \DesignVar_{\IterOpt+1}, \ObjFun^{(\LevFid_{\IterOpt+1})}(\DesignVar_{\IterOpt+1}), \LevFid_{\IterOpt+1} \}$

\STATE $\IterOpt+1 \leftarrow \IterOpt$

\UNTIL {$\Budget_{\IterOpt} \leq \Budget_{max}$}


\STATE \textbf{return} $\MinDesignVar$ that minimize $\ObjFun(\DesignVar)$ over $\Dataset_{\IterOpt}$ 

\end{algorithmic}

\end{algorithm}

\vspace{1cm}
 
Algorithm \ref{a:OptimizationAlgorithm} illustrates the pseudo-code of the multifidelity Bayesian optimization methodology. MFBO consists of a sequential approach to solve the optimization problem in Equation \eqref{e: optproblem}, starting with an initial dataset $\Dataset_{1} = \{ \DesignVar_{\IndNumObs}, \ObjFun^{(\LevFid_{\IndNumObs})}(\DesignVar_{\IndNumObs}), \LevFid_{\IndNumObs} \}_{\IndNumObs=1}^{\NumObs_1} $ of samples and associated values of the objective function evaluated with the $\LevFid_{\IndNumObs}$ level of fidelity. At each iteration $\IterOpt$ of the optimization procedure, the collected dataset is used to build the multifidelity surrogate model combining the prior distribution $\Probability(\ObjFun^{(\LevFid)})$ with the likelihood function $\Probability(\Dataset_{\IterOpt} | \ObjFun^{(\LevFid)})$ to compute the posterior distribution $\Probability(\ObjFun^{(\LevFid)} | \Dataset_{\IterOpt}) \propto \Probability(\Dataset_{\IterOpt} | \ObjFun^{(\LevFid)}) \Probability(\ObjFun^{(\LevFid)})$, which represents the updated emulator of the objective function for each level of fidelity. Then, MFBO induces a multifidelity acquisition function $\AF(\DesignVar, \LevFid)$ based on the posterior that measures the utility of probing domain locations with certain levels of fidelity: the next design $\DesignVar_{\IterOpt+1}$ and level of fidelity $\LevFid_{\IterOpt+1}$ to query are selected by maximizing the acquisition function $\{\DesignVar_{\IterOpt+1},\LevFid_{\IterOpt+1} \} = \max \AF(\DesignVar, \LevFid)$ over the domain $\DesignSpace$. After the new observation is collected, the dataset $\Dataset_{\IterOpt+1} = \Dataset_{\IterOpt} \cup  \{ \DesignVar_{\IterOpt+1}, \ObjFun^{(\LevFid_{\IterOpt+1})}(\DesignVar_{\IterOpt+1}), \LevFid_{\IterOpt+1} \}$ is updated and the procedure iterates until a maximum computational budget $\Budget_{\IterOpt} = \Budget_{max}$ is reached, where $\Budget_{\IterOpt}$ is the cumulative computational cost expended until iteration $\IterOpt$.

\section{Physics-Aware Multifidelity Bayesian Optimization}
\label{s: Methodology}

The Physics-Aware Multifidelity Bayesian Optimization (PA-MFBO) framework permits to accelerate the optimization procedure alleviating the burden of learning the characterization of the domain structure entirely from data. In particular, PA-MFBO embeds the available prior scientific and expert knowledge about the structure of the physical domain during the search procedure. This formalizes a sort of physics-aware reasoning based on an original formulation of the multifidelity acquisition function (Section \ref{s:DA-MFAF}) that biases the query of numerical models according to the specific structures of the domain, while still informed with data from the multifidelity emulator (Section \ref{s:MFGP}).

\subsection{Multifidelity Gaussian Process}
\label{s:MFGP}

\newcommand{\RegressParam}{\beta}
\newcommand{\ProcessVar}{\varsigma}
\newcommand{\RegressFun}{\upsilon}
\newcommand{\RoughParam}{\varpi}

Our PA-MFBO framework employs the multifidelity Gaussian process as the emulator of the objective function, which is formalized extending the Gaussian process to multiple levels of fidelity through an autoregressive scheme. 

The Gaussian process (GP) is a non-parametric kernel-based statistical model that permits to effectively approximate the black-box relationship between locations of the domain $\DesignVar$ and the associated values of the objective $\ObjFun(\DesignVar)$ \cite{williams1995gaussian, schulz2018tutorial}. Accordingly, the Gaussian process regression builds an emulator of the objective function using the knowledge of observations of $\ObjFun(\DesignVar)$ collected in specific locations, and defines a distribution over functions completely specified by the mean function $\MeanFunGP(\DesignVar) : \DesignSpace \rightarrow \mathbb{R}$ and covariance function $\CovFunGP(\DesignVar, \DesignVar') : \DesignSpace \times \DesignSpace \rightarrow \mathbb{R}$. The mean function $\MeanFunGP(\DesignVar) = \Expectation \left[ \ObjFun(\DesignVar) \right]$ reflects the expected value of the objective at a certain location $\DesignVar$ and the covariance or kernel function $\CovFunGP(\DesignVar, \DesignVar') = \Expectation \left[ (\ObjFun(\DesignVar)-\MeanFunGP(\DesignVar))(\ObjFun(\DesignVar')-\MeanFunGP(\DesignVar')) \right]$ represents the dependence between the values of the objective function at different locations $\DesignVar$ and $\DesignVar'$. This constitutes a predictive framework where the predictor in the form of the mean function approximates the objective function over unknown regions of the domain, and the covariance function quantifies the uncertainty associated with this prediction.

The availability of multiple representations of the objective function $\{\ObjFun^{(\LevFid)}\}_{\LevFid=1}^{\LevFid = \MaxLevFid}$ demands for an emulator capable to approximate the distribution of the objective function and synthesize observations from different sources of information. In this scenario, the Gaussian process regression can be extended to combine the models of the objective function at different levels of fidelity into a single predictive framework. Let us assume we have collected paired input/output observations of the objective function in the form $\Dataset_{\NumObs}=\{ \DesignVar_{\IndNumObs}, \NoisyObserv^{(\LevFid_{\IndNumObs})}(\DesignVar_{\IndNumObs}), \LevFid_{\IndNumObs} \}_{\IndNumObs=1}^{\NumObs}$, where the output $\mathbi{\NoisyObserv}=\{ \NoisyObserv^{(\LevFid_{\IndNumObs})}(\DesignVar_{\IndNumObs}) \}_{\IndNumObs=1}^{\NumObs}$ is normally distributed given $\mathbi{\ObjFun}=\{\ObjFun^{(\LevFid_{\IndNumObs})}_{\IndNumObs}\}_{\IndNumObs=1}^{\NumObs}$: 

\begin{equation}\label{e:MFGP0}
    \mathbi{\NoisyObserv} \; \vert \; \mathbi{\ObjFun}, \StandDevNoise^2 \sim \NorDist(\mathbi{\ObjFun},\StandDevNoise^2 \IdMatrix)
\end{equation}

\noindent where the measurement noise is characterized by the same variance $\StandDevNoise^2$ over the available levels of fidelity. 

Following the Bayesian inference principle, the prior belief about the objective function $\Probability(\ObjFun^{(\LevFid)})$ is combined with the likelihood function $\Probability(\Dataset_{\NumObs} \; \vert \; \ObjFun^{(\LevFid)})$ to compute the posterior distribution $\Probability( \ObjFun^{(\LevFid)}  \; \vert \; \Dataset_{\NumObs})  \propto \Probability(\Dataset_{\NumObs} \; \vert \; \ObjFun^{(\LevFid)}) \Probability(\ObjFun^{(\LevFid)})$. This represents the updated probabilistic model of the objective function at a certain level of fidelity $\LevFid$. In the black-box setting, the multifidelity Gaussian process regression considers no prior information about the multiple representations objective function. Hence, the lowest-fidelity prior $\ObjFun^{(1)} \sim GP(0,\CovFunGP_{1} \left( \DesignVar, \DesignVar' \right))$ is represented as a Gaussian process with zero mean function $\MeanFunGP^{(\LevFid)} = 0$ and covariance function $\CovFunGP_{1} \left( \DesignVar, \DesignVar' \right)$ while the higher levels of fidelity are derived recursively through an autoregressive scheme \cite{kennedy2000predicting}:

\begin{equation}\label{e:MFGP1}
\ObjFun^{(\LevFid)} = \ConsFact^{(\LevFid-1)} \ObjFun^{(\LevFid - 1)} \left( \DesignVar \right) + \Discrep^{(\LevFid)} \left( \DesignVar \right) \quad \LevFid = 2,...,\MaxLevFid
\end{equation} 

\noindent where $\ConsFact^{(\LevFid-1)}$ is a constant scaling factor that tunes successive levels of fidelity $\ObjFun^{(\LevFid)}$ and $\ObjFun^{(\LevFid - 1)}$, and $\Discrep^{(\LevFid)}$ is the discrepancy between two adjoining levels of fidelity modeled as a Gaussian process with mean function $\RegressFun(\DesignVar)^{T} \RegressParam^{(\LevFid)}$ and covariance function $\CovFunGP^{(\LevFid)} \left( \DesignVar, \DesignVar'\right)$ where $\RegressFun$ is the vector of regression functions and $\RegressParam^{(\LevFid)}$ are the regression coefficients. In this work, we adopt the Gaussian correlation model as the covariance function:

\begin{equation} \label{eq:kernelfunc} 
    \CovFunGP(\DesignVar, \DesignVar') = \ProcessVar^2_{\LevFid} \exp \{ - \sum_{m=1}^{M} \RoughParam^m_{\LevFid} (\DesignVar_m - \DesignVar_m')^2  \}
\end{equation}

\noindent where $\RoughParam = (\RoughParam^1_{\LevFid}, \RoughParam^2_{\LevFid}, ..., \RoughParam^M_{\LevFid})$ is the roughness parameter, and $\ProcessVar^2_{\LevFid}$ is the process variance of the $\LevFid$-th level of fidelity.

Accordingly, the posterior distribution of the multifidelity Gaussian process is completely specified by the mean $\MeanFunGP^{(\LevFid)}$ and variance $\StandDevGP^{2(\LevFid)}$: 

\begin{equation}\label{e:MFGP2}
\MeanFunGP^{(\LevFid)} (\DesignVar) = \CovFunGP_\NumObs^{(\LevFid)} (\DesignVar)^T \left(\KernelMatrix + \StandDevNoise \mathbi{I} \right)^{-1} \mathbi{\NoisyObserv}
\end{equation} 

\begin{equation}\label{e:MFGP3}
\StandDevGP^{2(\LevFid)} (\DesignVar) = \CovFunGP \left( \left(\DesignVar, \LevFid\right), \left( \DesignVar, \LevFid \right) \right) - \CovFunGP_\NumObs^{(\LevFid)} (\DesignVar)^T \left(\KernelMatrix + \StandDevNoise \mathbi{I} \right)^{-1} \CovFunGP_\NumObs^{(\LevFid)} (\DesignVar)
\end{equation}

\noindent where $\CovFunGP_\NumObs^{(\LevFid)}$ is defined as $ \CovFunGP_\NumObs (\DesignVar) \doteq \left( \CovFunGP \left( \left(\DesignVar, \LevFid \right), \left( \DesignVar_{1}, \LevFid_{1} \right) \right), \cdots , \CovFunGP \left(\left(\DesignVar, \LevFid \right), \left( \DesignVar_{\NumObs}, \LevFid_{\NumObs} \right) \right) \right)$, and $ \KernelMatrix$ is the kernel matrix formalized as follows: 

\begin{equation} \label{e:MFGP4}
    \KernelMatrix = \begin{pmatrix}
    \CovFunGP^{(\LevFid-1)}(\DesignVar, \DesignVar') \KernelMatrix^{(\LevFid-1)} & \ConsFact\CovFunGP^{(\LevFid-1)}(\DesignVar, \DesignVar')\KernelMatrix^{(\LevFid-1)} \\
    \ConsFact\CovFunGP^{(\LevFid-1)}(\DesignVar, \DesignVar')\KernelMatrix^{(\LevFid-1)} & \ConsFact^2\CovFunGP^{(\LevFid-1)}(\DesignVar, \DesignVar')\KernelMatrix^{(\LevFid-1)} + \CovFunGP^{(\LevFid)}(\DesignVar, \DesignVar')\KernelMatrix^{(\LevFid)}
    \end{pmatrix}
\end{equation}

\noindent where $\KernelMatrix^{(\LevFid-1)}(i,j) = \CovFunGP \left( \left( \DesignVar_{i}, \LevFid-1 \right), \left( \DesignVar_{j}, \LevFid -1 \right) \right)$ and $\KernelMatrix^{(\LevFid)}(i,j) = \CovFunGP \left( \left( \DesignVar_{i}, \LevFid \right), \left( \DesignVar_{j}, \LevFid \right) \right)$.

The posterior mean function $\MeanFunGP^{(\LevFid)}$ constitutes the prediction of the objective function at the $\LevFid$-th level of fidelity over the domain $\DesignSpace$, and the posterior standard deviation $\StandDevGP^{(\LevFid)}$ quantifies the associated level of uncertainty. The hyperparameters $(\ConsFact, \RegressParam, \RoughParam, \ProcessVar)$ of the multifidelity Gaussian process surrogate model are estimated by maximizing the likelihood function \cite{ForresterAl2008}.

\subsection{Physics-Aware Multifidelity Acquisition Function}
\label{s:DA-MFAF}

PA-MFBO incorporates the prior scientific and expert knowledge about the physical phenomena and the specific structure of the domain through the original physics-aware multifidelity acquisition function $\AF_{PA}(\DesignVar, \LevFid)$ based on the multifidelity expected improvement \cite{huang2006sequential}:

\begin{equation} \label{e:MFAF1}
\AF_{PA}(\DesignVar, \LevFid) =  \AF_{EI}(\DesignVar) \alpha_1(\DesignVar,\LevFid) \alpha_2(\DesignVar,\LevFid)  \alpha_3(\LevFid)\alpha_4(\PhysicsVec,\LevFid )
\end{equation} 
\noindent where $\AF_{EI}(\DesignVar)$ is the expected improvement acquisition function evaluated at the highest level of fidelity \cite{jones1998efficient}: 

\begin{equation} \label{e: expimprov}
\AF_{\EI}(\DesignVar) =  \Expectation [\max(\ObjFun^{(\MaxLevFid)}(\hat{\DesignVar}^*) - \ObjFun^{(\MaxLevFid)}(\DesignVar),0)] = 
     \StandDevGP(\DesignVar)(\Improv(\DesignVar)\CumDistFun(\Improv(\DesignVar))) + \NorDist(\Improv(\DesignVar); 0,1)
\end{equation}

\noindent where $\Improv(\DesignVar) = (\ObjFun^{(\MaxLevFid)}(\hat{\DesignVar}^*) - \MeanFunGP(\DesignVar))/\StandDevGP(\DesignVar)$ is the predicted improvement, $\hat{\DesignVar}^*$ is the current location of the best value of the objective sampled so far, $\CumDistFun(\cdot)$ is the cumulative distribution function of a standard normal distribution. The expectation term in Equation \ref{e: expimprov} relates only to the high-fidelity model of the objective function, and quantifies the expected gain potentially achieved adding an high-fidelity evaluation of the objective function. 

The terms $\alpha_1$, $\alpha_2$ and $\alpha_3$ are conceived to capture and balance the contributions of lower-fidelity evaluations of the objective function, and are formalized as follows:

\begin{equation}\label{e:MFAF3}
    \alpha_1 (\DesignVar, \LevFid) = corr \left[ \ObjFun^{(\LevFid)}(\DesignVar), \ObjFun^{(\MaxLevFid)}(\DesignVar) \right] = \frac{\CovFunGP((\DesignVar,\LevFid),(\DesignVar, \MaxLevFid))}{\sqrt{\StandDevGP^{2(\LevFid)}\StandDevGP^{2(\MaxLevFid)}}}
\end{equation}  
\begin{equation} \label{e:MFAF4}
    \alpha_2 (\DesignVar, \LevFid) = 1 - \frac{\StandDevNoise}{\sqrt{\StandDevGP^{2(\LevFid)} (\DesignVar) + \StandDevNoise^{2}}}
\end{equation}
\begin{equation} \label{e:MFAF5} 
    \alpha_3 (\LevFid) = \frac{\CompCost^{(\MaxLevFid)}}{\CompCost^{(\LevFid)}}.
\end{equation}

\noindent $\alpha_1$ is defined as the posterior correlation coefficient between the $\LevFid$-th level of fidelity and the highest-fidelity available at the same location of the domain. This utility function reflects the reduction of the acquisition function when samples are evaluated with lower-fidelity models, and accounts for the decrease of the accuracy associated with a low-fidelity representation of the objective function. Accordingly, the use of a high-fidelity model is solicited when a low-fidelity estimate might produce unreliable observations of the objective function. $\alpha_2$ considers the reduction of the uncertainty associated with the prediction of the multifidelity Gaussian process at the $\LevFid$-th level of fidelity after a new observations of the objective function with a certain level of fidelity $\LevFid$ is added to the dataset $\Dataset_{\NumObs}$. The objective of this term is to adjust the contribution of the high-fidelity expected gain quantified through Equation \ref{e: expimprov} considering the reduction of the optimization gain of additional evaluations at the $\LevFid$-th level of fidelity as the MFGP prediction becomes more accurate. Accordingly, this prevents the systematic sampling in already explored regions of the domain characterized by lower uncertainty. $\alpha_3$ is formulated as the ratio between the computational cost $\CompCost^{(\MaxLevFid)}$ associated with the evaluation of the high-fidelity model and the computational cost $\CompCost^{(\LevFid)}$ required to compute the $\LevFid$-th fidelity model. This utility function is conceived to include awareness about the computational resources required for the evaluation of the objective function adopting the $\LevFid$-th level of fidelity. The purpose of this term is to privilege the selection of lower-fidelity queries when similar improvements of the solution are obtained from higher-fidelity observations, and balance the computational cost and the informative contribution of different fidelity levels.

$\alpha_4(\PhysicsVec,\LevFid)$ is the physics-aware utility function that embeds a source of prior knowledge represented by a set of physical variables $\PhysicsVec$ in the sampling scheme. This permits to introduce a learning bias that captures the scientific understanding and expertise underlying the physical domain of the system. Without claiming to limit the informative content that can be incorporated into our acquisition function, we identify two main sources of prior knowledge in the form of scientific and expert knowledge. Scientific knowledge refers to the body of rules formalized and validated through the scientific method such as conservation laws, physical principles or phenomenological behaviors that represent the physics of interest. Examples include the Navier-Stokes partial differential equations in fluid dynamics to model the motion of viscous fluids, and numerical methodologies to approximate the solution of the governing equations as the finite element method in structural mechanics. Expert knowledge represents the information that is held by a community of experienced specialists and validated implicitly over several years of experience in a specific field. Examples include the common knowledge within the engineering or physics community resulting from training, research and personal experience. 

The proposed physics-aware multifidelity Bayesian optimization framework is illustrated and demonstrated for two cases of study, namely an aerodynamic design optimization problem (Section \ref{s: AeroOptimization}), and a structural health monitoring task (Section \ref{s: structprob}).

\section{Aerodynamic Design Example}
\label{s: AeroOptimization}

The design test case addresses the optimization of a transonic airfoil to improve the aerodynamic performance. Particular attention is dedicated to the cross-regime scenario where the fluid regime and the associated physical phenomena evolve during the optimization process. This defines a robust optimization procedure that potentially ensures to obtain optimal airfoil shapes for different operational conditions, without limiting the improvement of performance to a single application context  \cite{drela1998pros,li2002robust}. In this case, the prior scientific knowledge about the structure of the domain relates to the transition of the fluid dynamic regimes during the optimization. Thus, the physics-aware utility function is formalized to bias the search procedure according to the evolution of the physical domain.

The Mach number $\Mach$ is the main physical variable that captures the evolution of the fluid domain, and constitutes a measure of the compressibility effects that modify the fluid structure. According to the fluid mechanics theory \cite{elger2020engineering, young2010brief}, the flow-field around streamlined bodies is defined subsonic for values of the Mach number lower than 0.8, and represents a condition characterized by the absence of discontinuities and the fluid properties vary continuously. As the Mach number approaches the sonic condition, discontinuities in the form of local shock waves start to appear in the fluid domain with the consequent separation of the viscous boundary layer. This mixed subsonic-supersonic flow field emerges for values of the Mach number between 0.8 and 1.2, and is commonly referred as the transonic regime. The interactions between shock waves and boundary layer determine an increase of the drag force, and unsteady effects generated by a shift of the center of pressure of the aerodynamic body. Therefore, the cross-regime scenario poses significant challenges associated with the modeling of complex physics. On one hand, the subsonic regime can be represented adopting simplifications in the aerodynamic modeling due to the smooth evolution of the flow field; on the other hand, the transonic regime requires the implementation of accurate and robust modeling techniques to capture non-linear phenomena in the unsteady mixed subsonic-supersonic flow.

\subsection{Optimization Problem: Cross-Regime Airfoil Design}

The aerodynamic design optimization problem consists in the identification of the optimal combination of design parameters that minimizes the drag coefficient $\DragCoef$ of a transonic airfoil, subject to a variety of aerodynamic and geometric constraints. For this demonstrative test-case, we adopt the RAE 2822 transonic airfoil that is modified through the code WG2AER developed by \cite{quagliarella2020open}, where the original shape of the airfoil is linearly combined with weighted shape modification functions to obtain new geometries. The aerodynamic design optimization problem is formulated as follows: 

\begin{subequations} \label{e:ProblemFormulation2}
\begin{align}
    \min_{\DesignVar \in \DesignSpace } \quad
    & \DragCoef(\DesignVar) \nonumber \\ 
    \label{e: ProblemFormulation2 : designavar}
    & \DesignVar = \left[\Weight_1,..., \Weight_6, \Mach \right] \\[5pt]
    \quad
     s.t. \quad
    \label{e: ProblemFormulation2 : Constraint1}
    & \LiftCoef = 0.824 \\[5pt]
    \quad
    \label{e: ProblemFormulation2 : Constraint2}
    & -0.1 \leq \PitchCoef \leq -0.01  \\[5pt]
    \quad
    \label{e: ProblemFormulation2 : Constraint3}
    & \Thick / \Chord = 0.1211 \\[5pt]
    \quad
    \label{e: ProblemFormulation2 : Constraint4}
    & \Radius \geq 0.007\Chord \\[5pt]
    \quad
    \label{e: ProblemFormulation2 : Constraint5}
    & \TrailAngle \geq 5^{\circ} \\[5pt]
    \quad
    \label{e: ProblemFormulation2 : Constraint6}
    & \Thick_{85}/\Chord \geq 0.02 \\[5pt]
    \quad
    \label{e: ProblemFormulation2 : MoveLimits}
    & \DesignSpace = \Interval_{\Weight} \times \Interval_{\Mach}
\end{align}
\end{subequations}

\noindent where the design parameters $\DesignVar = \left[\Weight_1,..., \Weight_6, \Mach \right]$ consist of six weights $\Weight_i$ assigned to the shape modification polynomial functions and the Mach number $\Mach$. The feasibility of the design configuration is subject to obtain certain aerodynamic performances in terms of lift coefficient $\LiftCoef$ and pitching momentum coefficient $\PitchCoef$, and the modified geometry must accomplish the constraints on the airfoil thickness $\Thick$, chord $\Chord$, trailing edge angle $\TrailAngle$, and thickness of the airfoil at the $85 \%$ of the chord $\Thick_{85}$. The search for optimal design configuration is limited to the domain $\DesignSpace$ bounded by the move limits imposed for the weights $\Interval_{\Weight} = \left[-1, 1\right]^{6}$ and for the Mach number $\Interval_{\Mach} = \left[0.6, 0.99\right]$. This allows for the exploration of different aerodynamic configurations, and improves the robustness of the optimization procedure in presence of an evolution of the fluid domain from the low subsonic to the transonic regime.

\subsection{Aerodynamic Models}

The fluid domain around the airfoil is modeled through the Reynolds Averaged Navier-Stokes (RANS) equations to capture the effects of turbulence that occur at high speed regimes. The differential formulation of RANS is mathematically expressed as follows:

\begin{equation} \label{e:HFmodel1}
    \Residuals \left( \ConsVar \right) = \frac{\partial \left( \ConsVar \right)}{\partial \Time} + \nabla \cdot \ConvFluxes - \nabla \cdot \ViscFluxes - \Source = 0 \quad in \; \CompDomain, \quad \Time > 0
\end{equation}

\noindent where $\CompDomain$ is the computational domain, $\Residuals$ are the numerical residuals, $\Source$ is the source term, $\ConsVar = \left( \Density, \Density \Velocity, \Density \Energy \right)$ are the conservative variables, namely the air density $\Density = \Density(\Altitude)$, the free-stream velocity $\Velocity$ and the total energy $\Energy$, and $\ConvFluxes$ and $\ViscFluxes$ are the convective and viscous fluxes, respectively: 

\begin{equation}
\ConvFluxes = \left(
\begin{array}{c}
      \Density \Velocity \\
       \Density \Velocity \otimes \Velocity + \mathbi{I} \Pressure \\
       \Density \Energy \Velocity +  \Pressure \Velocity 
\end{array} \right)
\end{equation}

\begin{equation}
\ViscFluxes = \left(
\begin{array}{c}
      \cdot \\
       \ViscStress \\
       \ViscStress \Velocity +  \Conductivity \nabla \Temperature      
\end{array} \right)
\end{equation}

\noindent where $\Temperature = \Temperature(\Altitude)$ is the free-stream temperature, $\Pressure = \Pressure(\Altitude)$ is the free-stream static pressure, $\Conductivity = \Conductivity(\Altitude)$ is the thermal conductivity and $\ViscStress$ is the tensor of viscous stresses. 

We are interested in the distribution of the pressure coefficient around the airfoil which is a function of the design configuration selected at each iteration of the optimization procedure. The aerodynamic modeling approach consists in the numerical solution of the RANS equations through a Computational Fluid Dynamic (CFD) solver to obtain a finite-dimensional approximation of the pressure coefficient over the fluid domain. We use the SU2 v6.2.0 CFD code based on the finite-volume method to discretize the RANS partial differential equations considering a fully turbulent flow-field \cite{economon2016su2}. To ensure the robustness of the aerodynamic outcomes, the convergence criteria is set for a computational residuals minor than $10^{-6}$ with a fixed maximum number of 20000 iterations. The fluid domain is discretized through a computational mesh generated using the GMSH software v4 \cite{geuzaine2009gmsh} where an automated procedure embedded within the adopted computational tool \cite{quagliarella2020open} adapts the hybrid grid of triangles and quadrangles elements with the modified geometry of the airfoil. This permits to target the balance between accuracy and efficiency of the CFD computations.

The pressure field around the airfoil is represented through three numerical models based on the aforementioned aerodynamic modeling approach, and differ for the accuracy and related CPU time associated with their evaluation. These models return the design objective $\DragCoef$ and the aerodynamic constraints on lift $\LiftCoef$ and pitching momentum coefficients $\PitchCoef$ given the selected design configuration in terms of modified geometry of the airfoil through the assignment of the weights $\mathbi{\Weight}$ and Mach number $\Mach$. The fidelity of the aerodynamic simulations is determined controlling the granularity of the computational mesh through the associated element scale factor $ES$ where the higher the value of $ES$ the coarser the discretization of the fluid domain. Specifically, three levels of fidelity are considered for the aerodynamic modeling: we set $ES=2.5$ for the high-fidelity model corresponding to a grid of about $90000$ cells, $ES=12$ for the mid-fidelity model with about $30000$ cells, and $ES=20$ for the low-fidelity model consisting of a mesh with $15000$ cells.

\begin{figure*}[t!]
    \centering
        \includegraphics[width=0.5\linewidth,trim=220 0 245 0, clip]{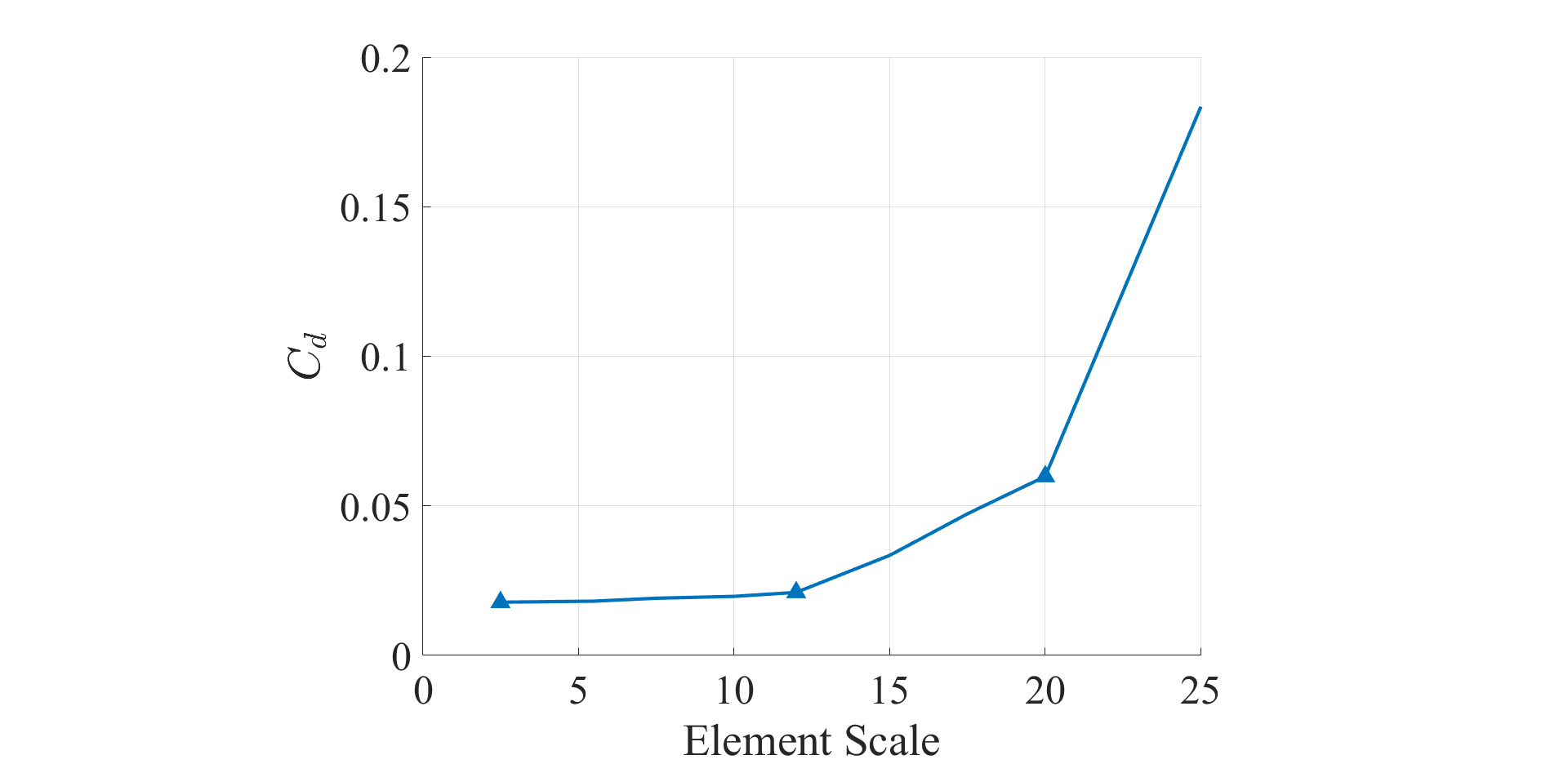} 
    \caption{Drag coefficient of the RAE-2822 airfoil varying the element scale of the CFD computational mesh. The markers indicate the drag coefficients computed adopting the element scale of the high-fidelity $ES=2.5$, mid-fidelity $ES=12$, and low-fidelity $ES=20$ aerodynamic model. }\label{fig:CFDconvergence}
\end{figure*}

Figure \ref{fig:CFDconvergence} illustrates the drag coefficient computed with the aerodynamic model for different element scale factors. The high-fidelity model -- marker corresponding to $ES=2.5$ in Figure \ref{fig:CFDconvergence} -- achieves an accurate representation of complex aerodynamic phenomena that occurs at higher regimes of speed including discontinuities, shock-waves and unsteadiness of the flow-field. This provides a close prediction of the mixed subsonic-supersonic fluid domain that characterizes the transonic regime. The mid-fidelity model -- marker corresponding to $ES=12$ in Figure \ref{fig:CFDconvergence} -- reduces the demand for CPU if compared with the high-fidelity model by decreasing the number of cells that discretize the fluid domain. This produces a reliable estimate of the aerodynamic coefficients for Mach number regimes far from the sonic condition where the unsteady phenomena have marginal effects, and a reduced accuracy for discontinuous flows that occur at the transonic regime. The low-fidelity model -- marker corresponding to $ES=20$ in Figure \ref{fig:CFDconvergence} -- further decreases the number of elements adopted to discretize the fluid domain, and leads to an inaccurate representation of the compressibility effects that characterize the more turbulent flows at higher Mach values ($\Mach>0.65$). However, the coarser discretization reduces the computational cost required for its evaluation of the $65\%$ with respect to the cost associated with the high-fidelity model.

\begin{figure*}[t!]
    \centering
     \subfigure[]{%
        \includegraphics[width=0.44\linewidth,trim=0cm 0 0cm 0, clip]{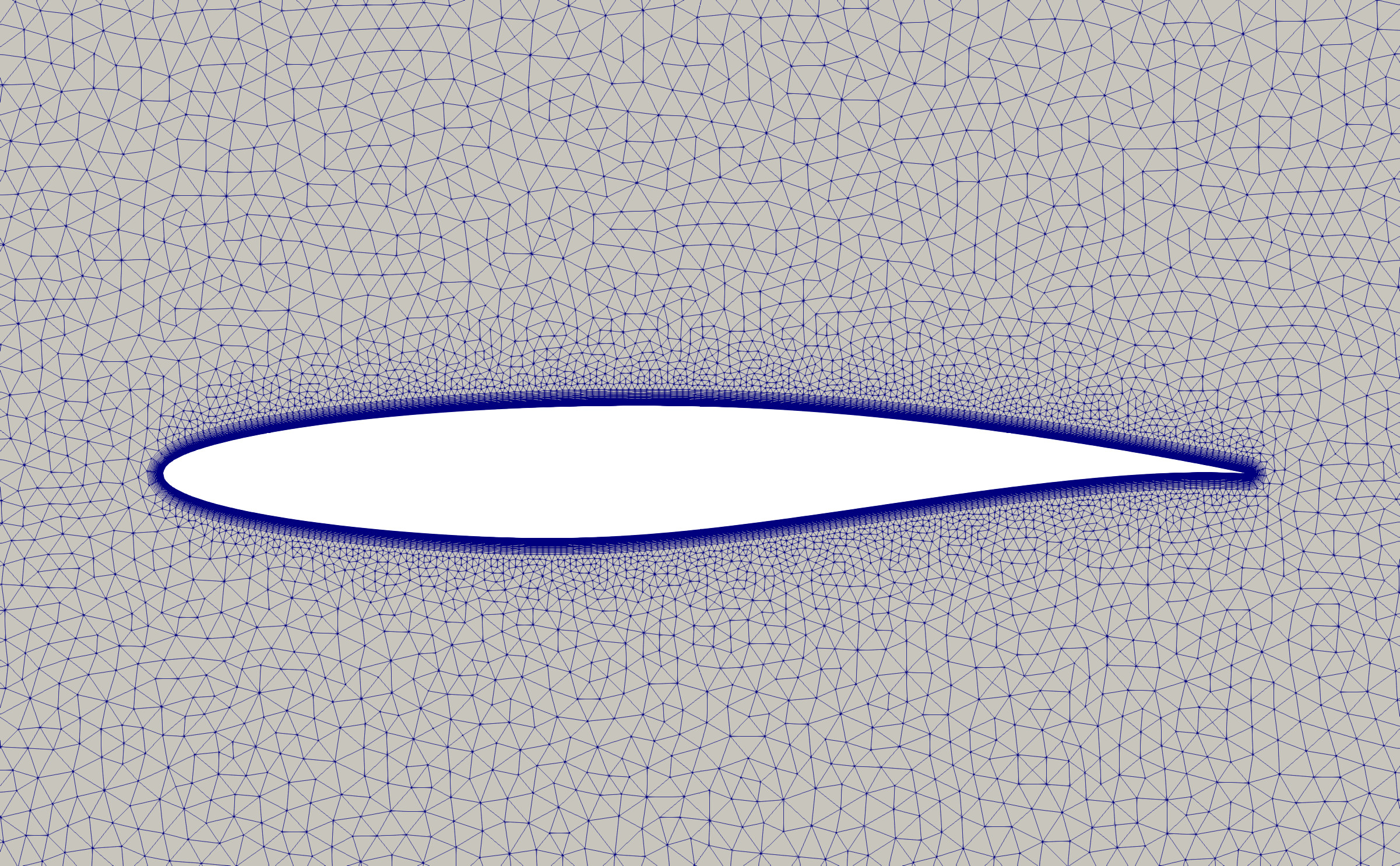} 
        }
        \subfigure[]{%
        \includegraphics[width=0.44\linewidth,trim=0cm 0 0cm 0, clip]{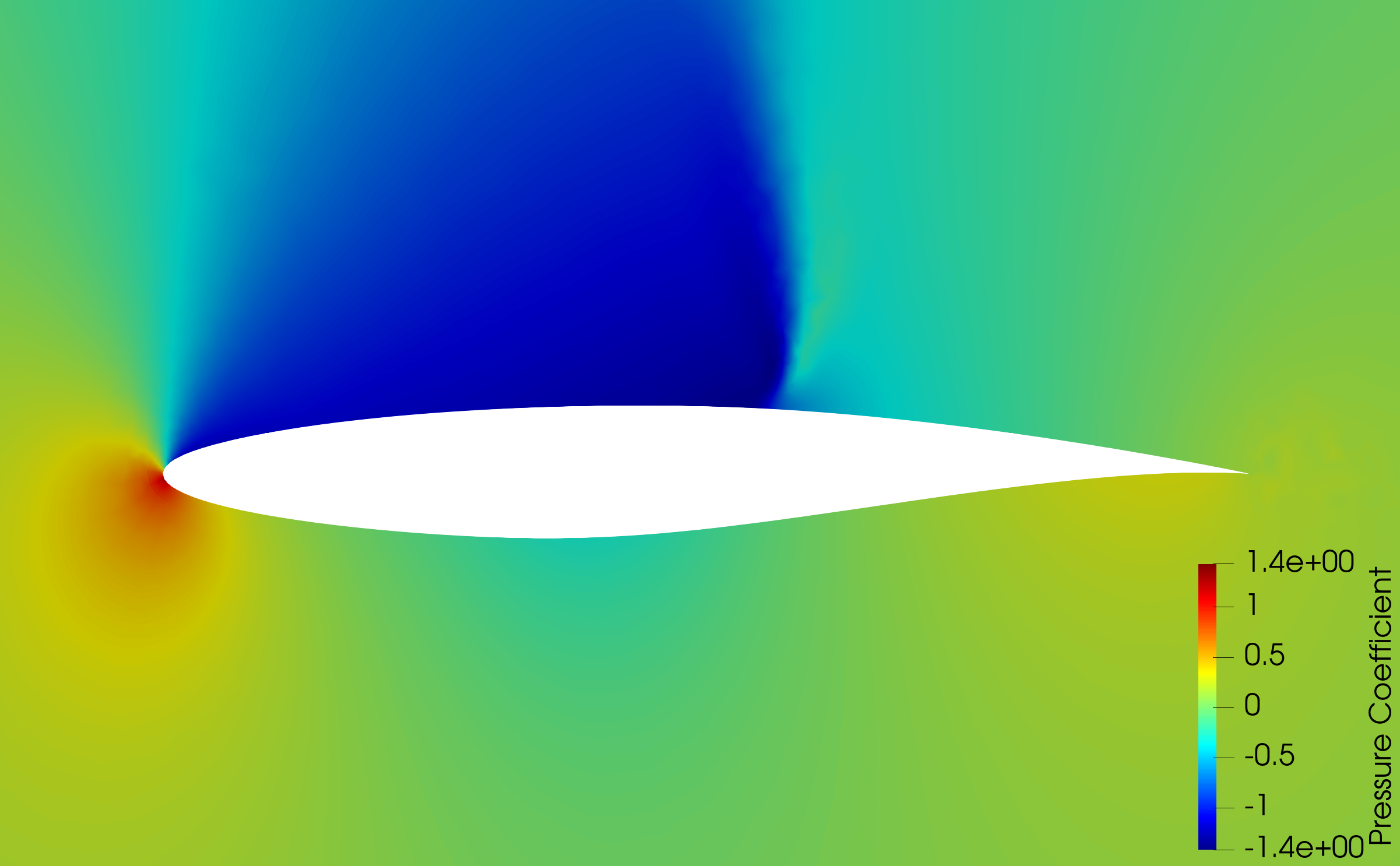}
        }

    \caption{(a) high-fidelity discretization of the computational domain, and (b) high-fidelity pressure coefficient contours for the RAE-2822 airfoil.}\label{fig:hfmodelaero}
\end{figure*}

\begin{figure*}[t!]
    \centering
     \subfigure[]{%
        \includegraphics[width=0.44\linewidth,trim=0cm 0 0cm 0, clip]{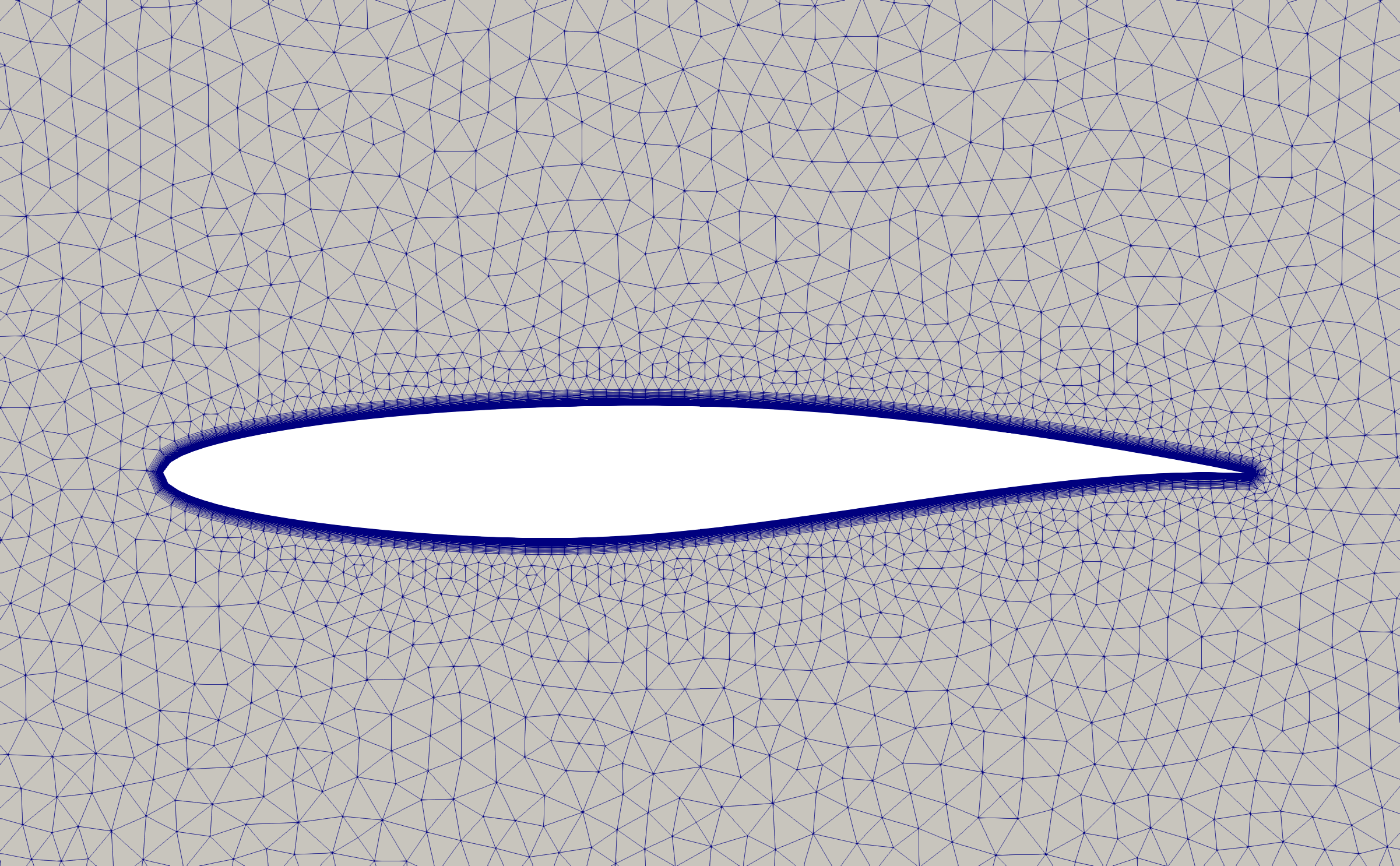} 
        }
        \subfigure[]{%
        \includegraphics[width=0.44\linewidth,trim=0cm 0 0cm 0, clip]{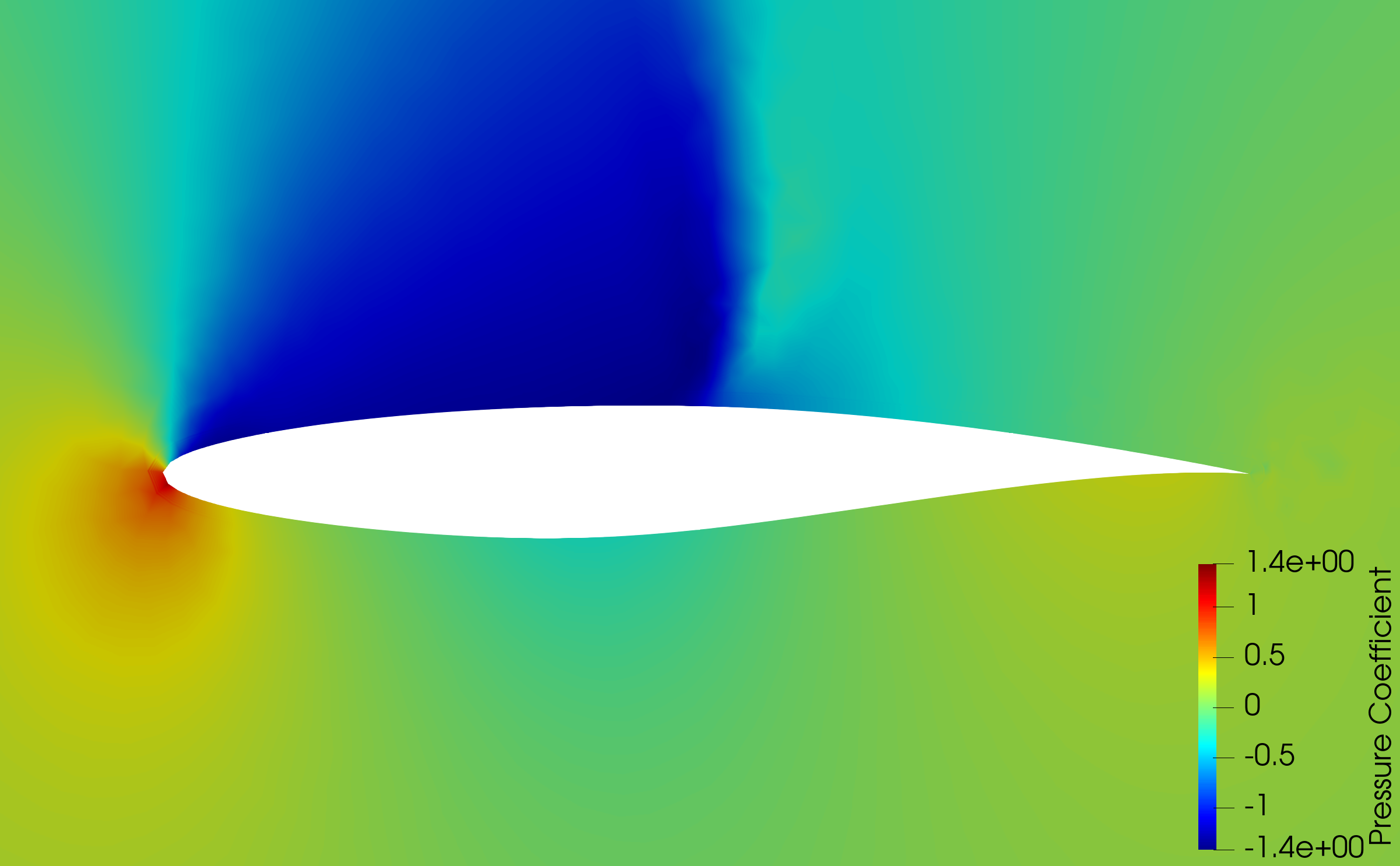}
        }

    \caption{(a) mid-fidelity discretization of the computational domain, and (b) mid-fidelity pressure coefficient contours for the RAE-2822 airfoil.}\label{fig:mfmodelaero}
\end{figure*}

\begin{figure*}[t!]
    \centering
     \subfigure[]{%
        \includegraphics[width=0.44\linewidth,trim=0cm 0 0cm 0, clip]{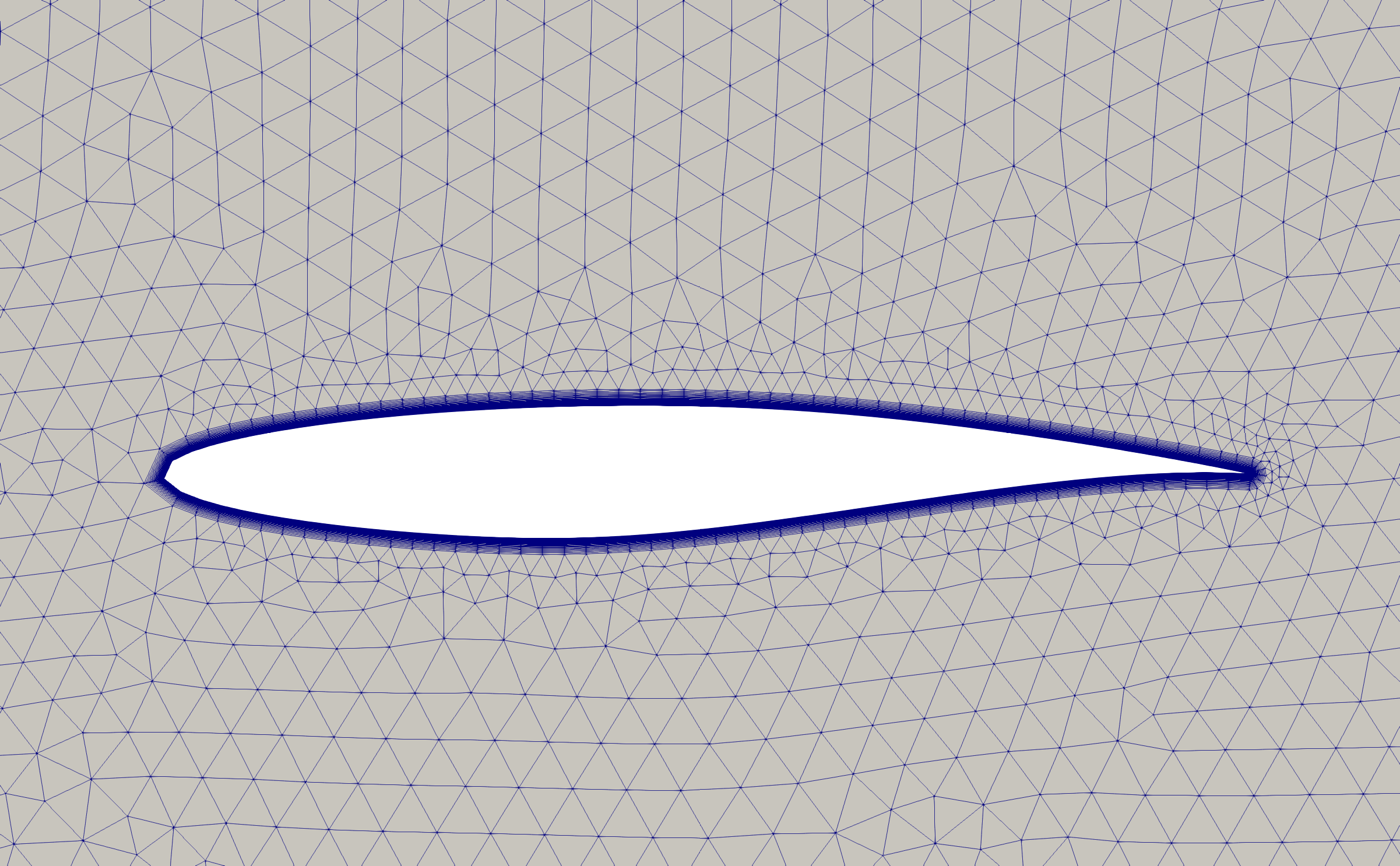} 
        }
        \subfigure[]{%
        \includegraphics[width=0.44\linewidth,trim=0cm 0 0cm 0, clip]{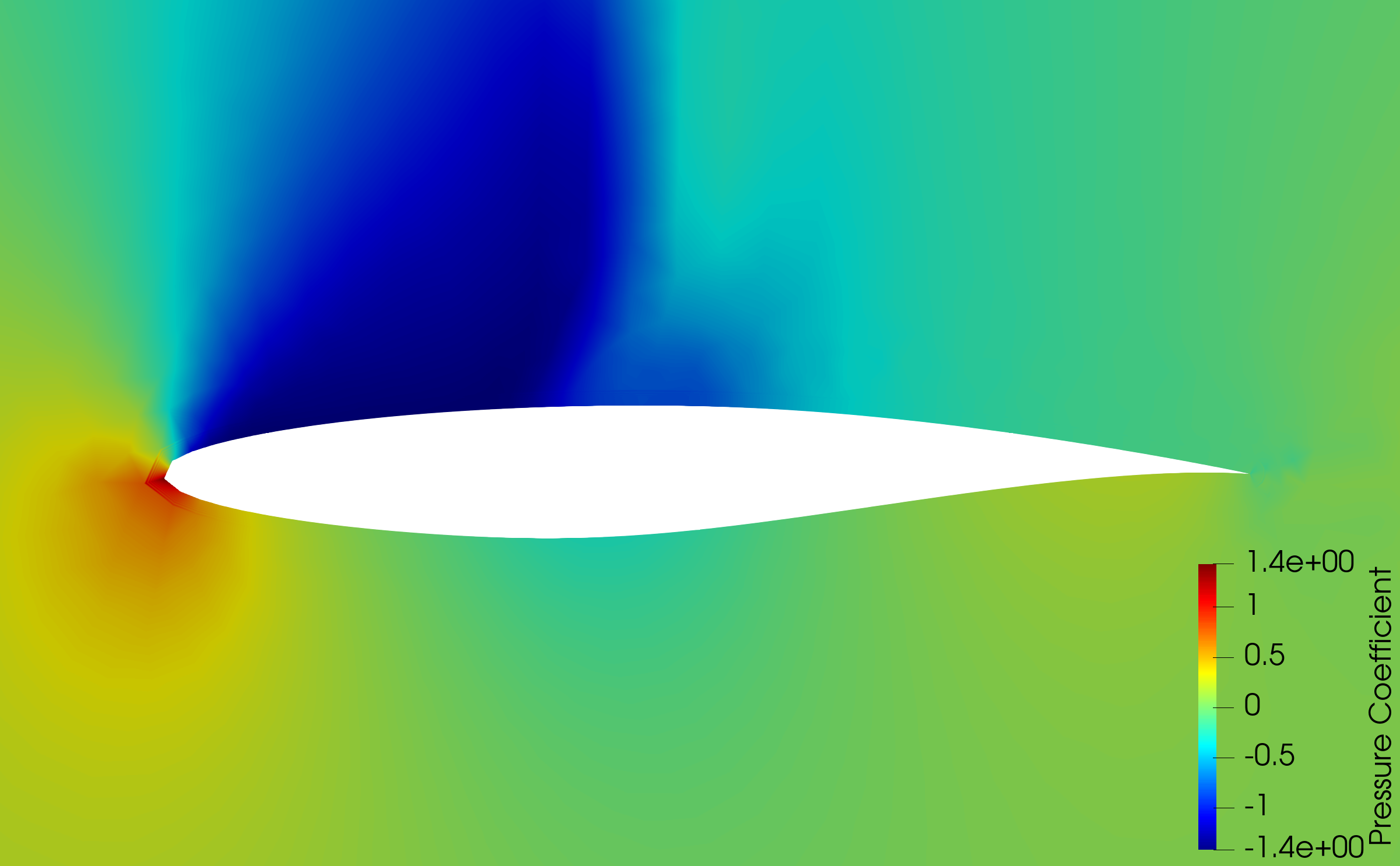}
        }

    \caption{(a) low-fidelity discretization of the computational domain, and (b) low-fidelity pressure coefficient contours for the RAE-2822 airfoil.}\label{fig:lfmodelaero}
\end{figure*}

For the aerodynamic representations, we define the relative computational cost $\CompCost^{(\LevFid)}$ of the $\LevFid$-th CFD model as the ratio between the element scale factors of the high-fidelity model and the $\LevFid$-th level of fidelity model. Accordingly, we set $\CompCost^{(\MaxLevFid)}=1$ for the high-fidelity aerodynamic model, $\CompCost^{(2)}=0.2$ for the mid-fidelity model, and $\CompCost^{(1)}=0.125$ for the low-fidelity model. The high-fidelity (Figure \ref{fig:hfmodelaero}), mid-fidelity (Figure \ref{fig:mfmodelaero}), and low-fidelity (Figure \ref{fig:lfmodelaero}) meshes and distributions of the pressure coefficient highlight the increasing discrepancy of the flow field representations as the level of fidelity decreases.

\subsection{Physics-Aware Utility Function for Aerodynamic Design}

We formulate our physics-aware utility function $\alpha_4$ to include a sort of physics-based reasoning that wisely quantifies the utility of querying an aerodynamic model according to the characteristics of the fluid dynamic regime. This is achieved by formalizing a bias dependent on the Mach number $\PhysicsVec = \Mach$ as the variable representative of the physical phenomena involved in the aerodynamic domain. Accordingly, $\alpha_4(\Mach, \LevFid)$ is formalized as follows:

\begin{equation} \label{e:MFAF6}
\begin{split}
\alpha_4 (\Mach, \LevFid) =
 & \left\{ \begin{array}{lll}  1 & \quad \mbox{if} & \LevFid = 1,...,\MaxLevFid-1  \\  \frac{\Mach_s}{\Mach_s - \Mach} & \quad \mbox{if} & \LevFid = \MaxLevFid  \quad \Mach_s = 1  \end{array} \right.
\end{split}
\end{equation}

This physics-aware utility function encourages the query of the high-fidelity model for values of the Mach number close to the sonic condition ($\Mach = 1)$. Indeed, $\alpha_4$ increases the value of the multifidelity acquisition function (Equation \eqref{e:MFAF1}) when an aerodynamic configuration is evaluated with the high-fidelity model in the transonic regime ($\Mach > 0.8)$. The goal is to capture large-scale separation of the fluid vein and unsteady effects that deeply influence the overall performance of the aerodynamic system. This permits to better support and improve the search for optimal designs through the a priori scientific knowledge about the aerodynamic domain structure derived directly from the governing equations.

\subsection{Aerodynamic Design Results}

This section illustrates and discusses the results achieved with the physics-aware multifidelity Bayesian optimization (PA-MFBO) framework for the aerodynamic design optimization problem of the RAE 2822 transonic airfoil. The effectiveness of the PA-MFBO algorithm is compared with other existing methods commonly adopted to address black-box optimization problems, namely the single-fidelity efficient global optimization (EGO) algorithm \cite{jones1998efficient}, and the multifidelity Bayesian optimization based on the multifidelity expected improvement acquisition function (MFBO) \cite{huang2006sequential}. For the aerodynamic design experiments, the initial set of samples is determined through a Latin hypercube strategy \cite{viana2016tutorial} and are used to compute the surrogate model at the first step (Section \ref{s:MFGP}). Specifically, the multifidelity algorithms are initialized with 32 initial aerodynamic design configurations among which 20 evaluations of the objective function are obtained with the low-fidelity model, 10 are computed with the mid-fidelity model, and 2 are evaluated with the high-fidelity model, while for the single fidelity frameworks we consider 6 initial design configurations at which we compute high-fidelity observations. We select the minimum drag coefficient as the assessment metric to evaluate the capabilities of the competing algorithms, and provide a measure of the improvement of the aerodynamic performance achieved by the identified design configurations:

\begin{equation} \label{e:mindrag}
    \DragCoef^{*} = \min (\DragCoef(\DesignVar))
\end{equation}

\begin{figure*}[t!]
    \centering

        \includegraphics[width=0.5\linewidth,trim=220 0 245 0, clip]{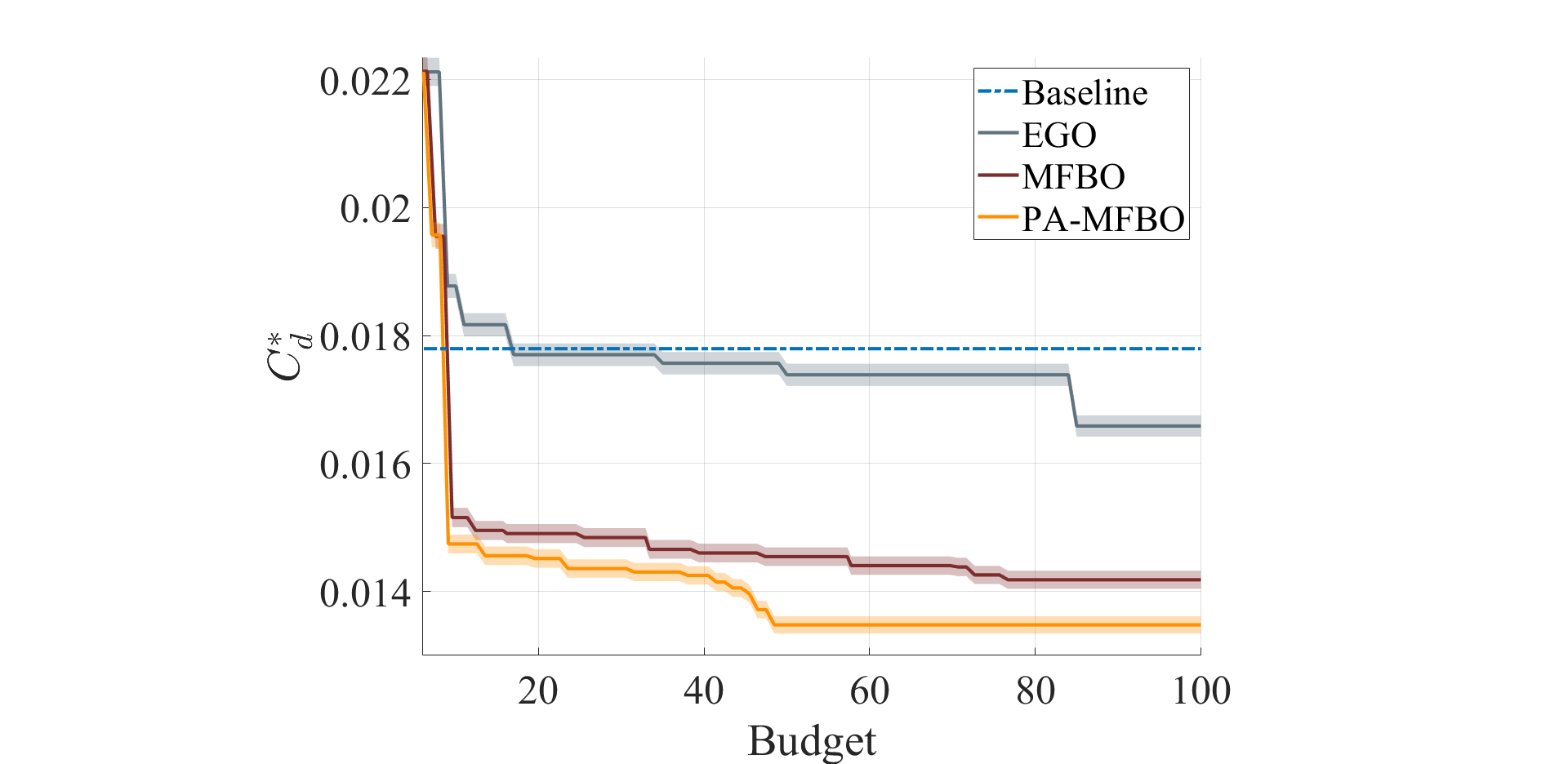} 
        \label{fig:CD}   
   
    \caption{Statistics over 25 runs of the minimum drag coefficient $\DragCoef^{*}$ obtained with the competing algorithms.}\label{fig:CDresults}
\end{figure*}

Figure \ref{fig:CDresults} reports the progression of the optimization procedure in terms of values of the minimum drag coefficient $\DragCoef^{*}$ as a function of the computational budget $\Budget$. We define the computational budget $\Budget = \sum \CompCost^{(\LevFid)}_{\IterOpt}$ as the cumulative computational cost $\CompCost^{(\LevFid)}_{\IterOpt}$ expended evaluating the $\LevFid$-th aerodynamic level of fidelity at each iteration of the optimization procedure. We compute 25 independent replications of the experiment for each methodology to measure and compensate the influence of the random initial sampling procedure and ensure a fair comparison of the algorithms. The outcomes of the statistics are represented through the median values of the assessment metric $\DragCoef^{*}$ together with the the associated values in between the 25-th and 75-th percentiles. We adopt as the baseline solution the value of the drag coefficient $\DragCoef^{*}=0.017796$ obtained for the RAE 2822 airfoil corresponding to the design configuration $\DesignVar = \left[0,0,0,0,0,0, 0.65 \right]$. The overall convergence histories show that all the algorithms are capable to identify promising design configurations, and provide significant reductions of the drag coefficient if compared with the baseline RAE 2822 design solution. However, the PA-MFBO method leads to superior design solutions in terms of aerodynamic performance if compared with EGO and MFBO, as a result of the better management and exploitation of multiple numerical models. As can be seen, PA-MFBO reduces the drag coefficient at the beginning of the optimization procedure and identifies optimal design configurations consuming a fraction of the available computational budget. In addition, we note from the convergence of the PA-MFBO experiments that the algorithm starts the search allocating budget for the exploration of different design configurations over the domain, which corresponds to a moderate reduction of the drag coefficient. Then, the computational resources are directed towards the exploitation phase reducing the values of the design objective. After the identification of an optimal design solution, it is possible to notice that PA-MFBO queries design solutions that perform worse that the best design identified so far -- the median of the minimum drag coefficient remains constant as the consumed budget increases. We observed that the algorithm in this phase mostly evaluates the low-fidelity aerodynamic model to contain the computational expense during a secondary exploration phase.

\begin{table} [t!]
\centering     
\begin{tabular}{lccc}
\hline\noalign{\smallskip}
$\Budget$ & $\DragCoef^{*}$ EGO & $\DragCoef^{*}$ MFBO & $\DragCoef^{*}$ PA-MFBO  \\
\noalign{\smallskip}\hline\noalign{\smallskip} 
6  & 0.02212 (-24.30 \%) &  0.02212 (-24.30 \%)& 0.02212 (-24.30 \%)\\
10  & 0.01887 (-6.055 \%)&  0.01515 (14.87 \%) & 0.01455 (18.24 \%)\\
25  & 0.01770 (0.5394 \%)&  0.01484 (16.61 \%) & 0.01435 (19.36 \%)\\
50 & 0.01738 (2.337 \%)  &  0.01454 (18.30 \%) & 0.01347 (24.31 \%)\\
100 & 0.01658 (6.833 \%) &  0.01418 (20.32 \%) & 0.01348 (24.31 \%)\\
\noalign{\smallskip}\hline
\end{tabular}
\caption{Median values of the minimum of the drag coefficient $\DragCoef^{*}$ and corresponding design improvement $(\cdot)$ obtained with the competing algorithms.}
\label{t:CDConvHist}
\end{table}

\begin{table} [t!]
\centering     
\begin{tabular}{lccc}
\hline\noalign{\smallskip}
Method & HF evaluations & MF evaluations & LF evaluations  \\
\noalign{\smallskip}\hline\noalign{\smallskip} 
EGO      & 100 & - & -  \\
MFBO     & 11.6 & 127.3 & 503.5  \\
PA-MFBO  & 12.3 & 99.5 & 542.4  \\
\noalign{\smallskip}\hline
\end{tabular}
\caption{Average evaluations of the high-fidelity (HF), mid-fidelity (MF), and low-fidelity (LF) aerodynamic model over the 25 runs of the aerodynamic design optimization problem.}
\label{t:Aero_levfidcalls}
\end{table}

To further quantify the performance of the PA-MFBO methodology, Table \ref{t:CDConvHist} reports the median values of the minimum drag coefficient for incremental computational expense $\Budget=$ 6, 10, 25, 50, and 100. At the end of the initial sampling phase ($\Budget=$ 6), the identified designs of all the algorithms determine values of the drag coefficient higher than the baseline solution. However, after the collection of data from the aerodynamic models ($\Budget=$ 10), the multifidelity frameworks are capable to improve the baseline design configuration while the EGO algorithm still achieves worst designs if compared with the unmodified RAE 2822 airfoil. The PA-MFBO methodology realizes the larger design improvement of the $24.31 \%$ before consuming a Budget of $\Budget=50$, which is superior to the MFBO design upgrade of the $20.32 \%$ obtained adopting much more computational resources. Moreover, the EGO methodology is capable to deliver a design improvement of only the $6.83 \%$ using all the available computational budget. Table \ref{t:Aero_levfidcalls} provides details about the average evaluations of the aerodynamic models at different levels of fidelity for each competing algorithm over the statistics of 25 runs. It is possible to observe that the multifidelity methods (MFBO and PA-MFBO) drastically reduce the acquisition of high-fidelity data with respect to the single-fidelity EGO algorithm, and use lower levels of fidelity to efficiently explore the design space. However, the proposed PA-MFBO permits to better direct the computational resources towards optimal design solutions through the wise evaluation of the costly high-fidelity model guided by the physics awareness about the evolution of the aerodynamic domain. This guarantees the identification of superior design solutions with contained computational cost if compared with the standard EGO and MFBO.

\begin{figure*}[t!]
    \centering
    
    \subfigure[]{%
        \includegraphics[width=0.48\linewidth,trim=100 0 30 0, clip]{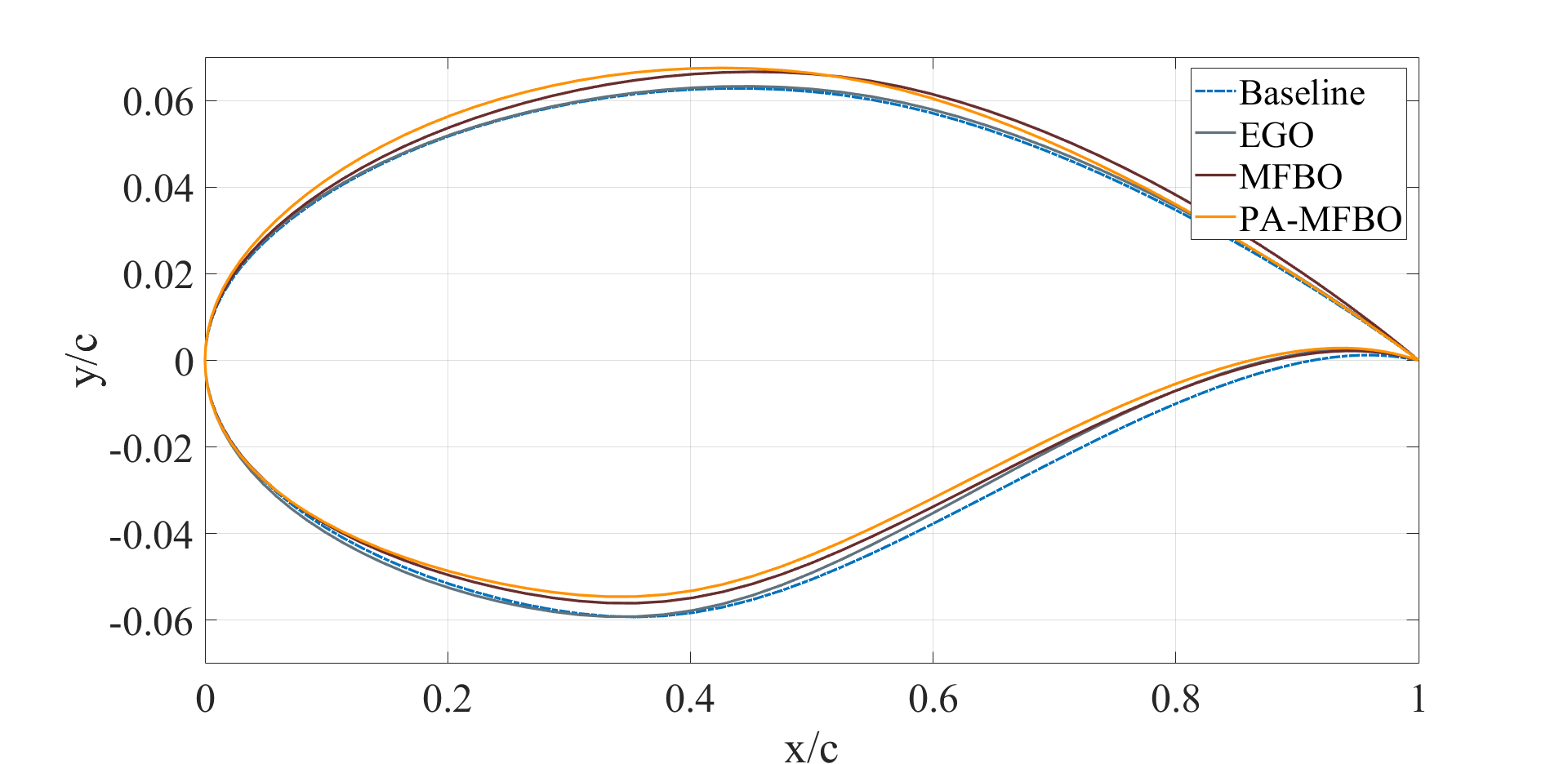}
        \label{fig:airfoil}}
     \subfigure[PA-MFBO]{%
        \includegraphics[width=0.42\linewidth,trim=0 0 0 0, clip]{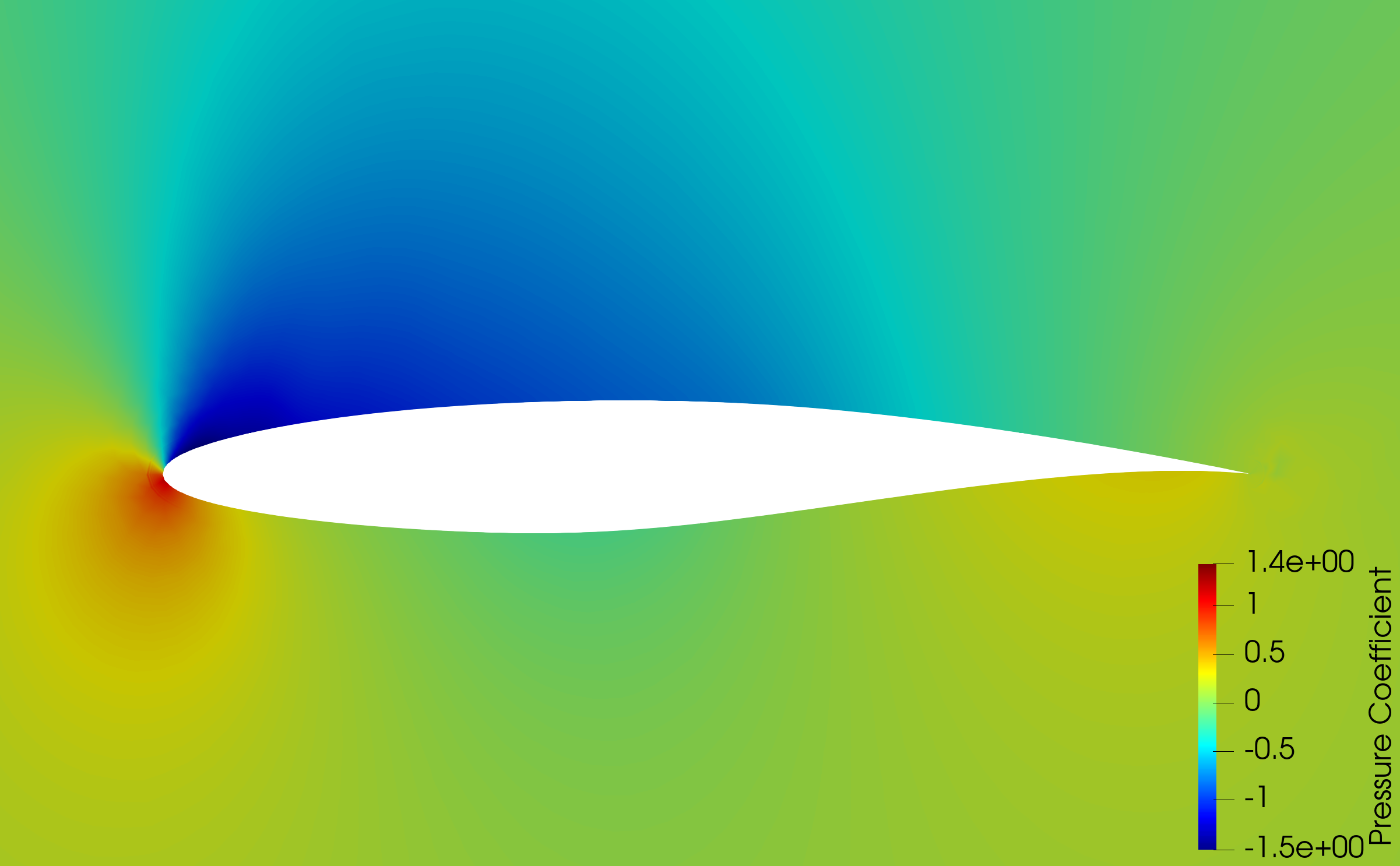} 
        \label{fig:pamfbo}}
        
    \subfigure[MFBO]{%
        \includegraphics[width=0.42\linewidth,trim=0 0 0 0, clip]{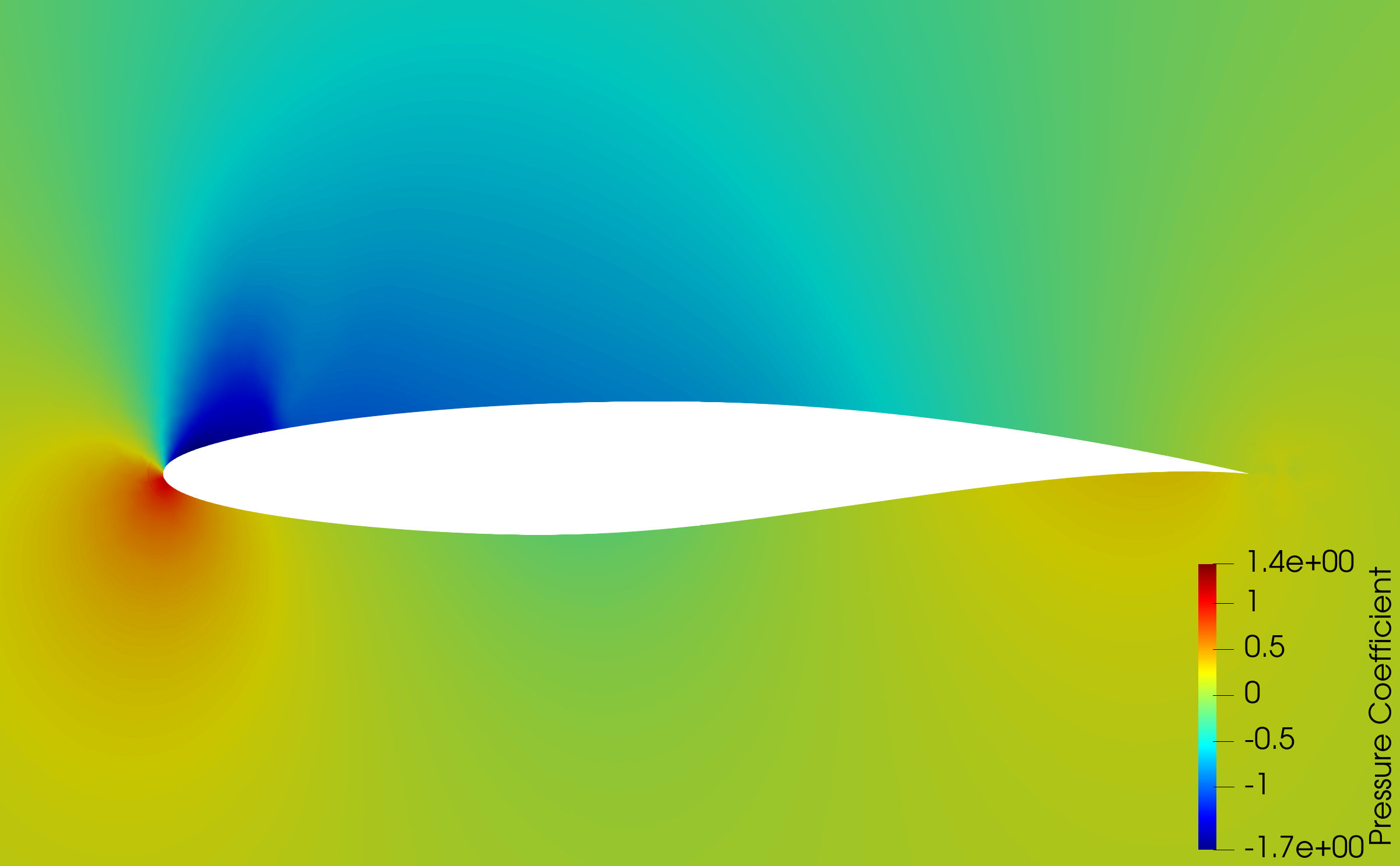}
        \label{fig:mfbo}}
        \hspace{0.85cm}
    \subfigure[EGO]{%
        \includegraphics[width=0.42\linewidth,trim=0 0 0 0, clip]{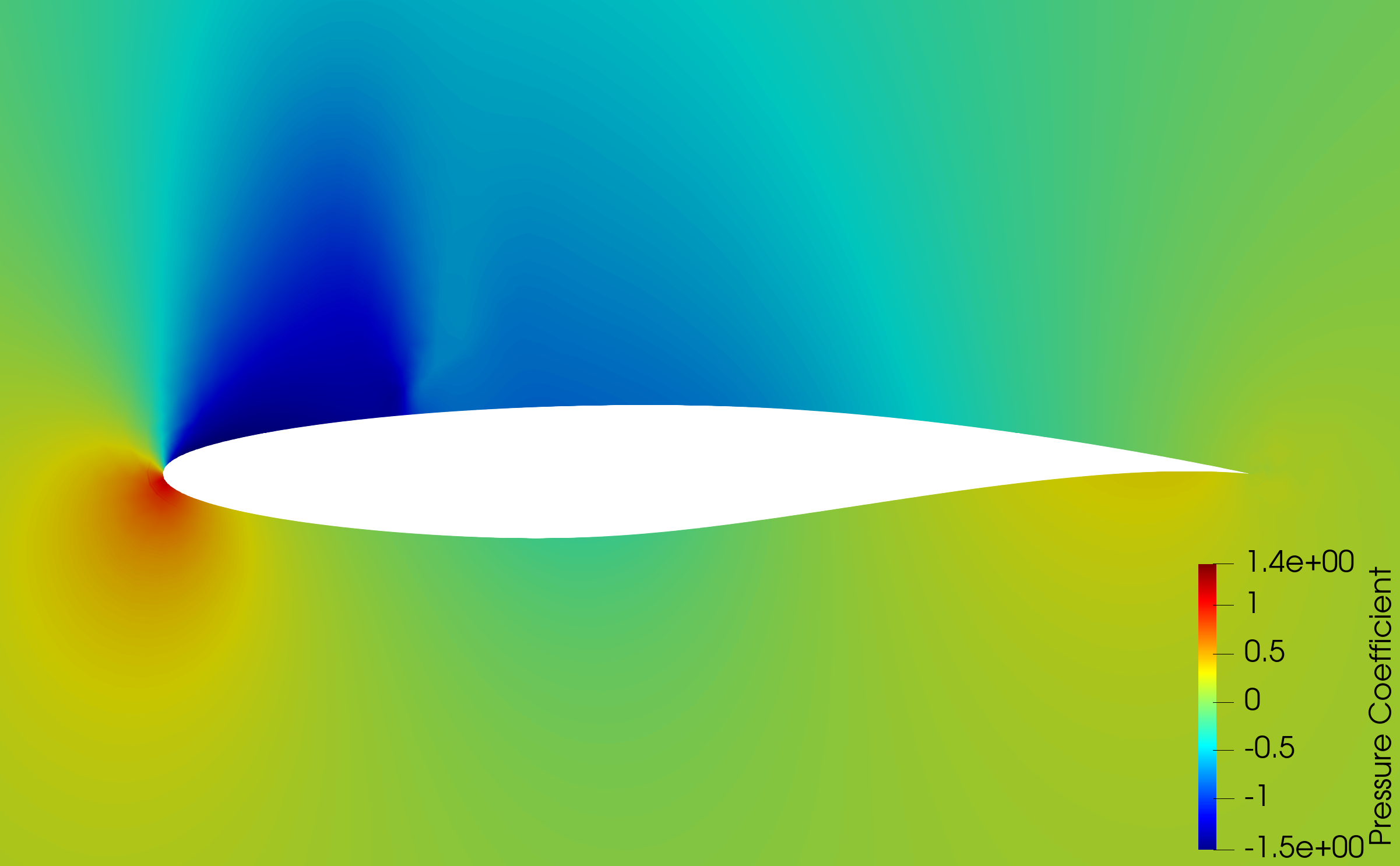}
        \label{fig:ego}}

    \caption{(a) Optimal airfoils geometry and associated pressure coefficient contours obtained with (b) the PA-MFBO, (c) MFBO, and (d) EGO algorithm.}\label{fig:cpoutcomes}
\end{figure*}

To clarify and interpret the results obtained, Figure \ref{fig:cpoutcomes} illustrates the aerodynamic performance of the optimal designs determined by all the algorithm. In particular, we report the optimal airfoil shapes corresponding to the best aerodynamic design (Figure \ref{fig:airfoil}), and the related pressure coefficient distribution for the PA-MFBO (Figure \ref{fig:pamfbo}), MFBO (Figure \ref{fig:mfbo}), and EGO (Figure \ref{fig:ego}) design solutions. It can be noticed that the superior performance of the design configuration identified with the PA-MFBO can be explained with the efficient expansion of the fluid at the upper surface leading edge that induces low-intensity shock waves if compared with the other modified airfoils. This results from the increase of the leading edge radius and aft camber that produces a reduction of the adverse pressure gradient, and permits a smooth evolution of the pressure coefficient in the supersonic bubble. This features of the aerodynamic domain determine the substantial decrease of the drag coefficient and enhance the overall efficiency of the modified airfoil. In this design test case, the remarkable performance of the PA-MFBO framework is related to the physical bias introduced in the sampling scheme. This enables the capitalization from the prior scientific knowledge about the fluid dynamic regime, and permits to accelerate and improve the optimization search through the wise selection of the aerodynamic model to query with a continuous balance between computational cost and accuracy of the solution.

\section{Structural Health Monitoring Example}
\label{s: structprob}

The structural health monitoring problem requires the assessment of the health status of a composite skin plate of an aircraft wing. Particular attention is dedicated to the incipient fracture of the carbon fiber: this represents one of the most critical failure for laminates since involves the degradation of the mechanical properties of the material and cannot be easily detected by standard non-destructive health monitoring techniques \cite{giurgiutiu2015structural}. For this application, the expert knowledge about the physics relates to specific structures of the domain characterized by damage conditions that might be misinterpreted by simplified modeling approaches. This knowledge is included in the search procedure through a physics-aware utility function that biases the query of numerical models to accurately distinguish the actual fault condition affecting the plate.

\subsection{Optimization Problem: Composite Plate Health Monitoring}
\label{s:healthmonitoringprob}

The structural health monitoring problem demands for the identification of the damage parameters affecting a composite plate subject to a cut in the fibers. The composite plate is constituted of four layers of plain weave fabric of carbon prepreg (IM7/8552 AS4) laminated with a stacking sequence $\left[ 45^{\circ} / 0^{\circ} / 0^{\circ} / 45^{\circ} \right]$, and with dimension of $102$ mm transversal length, $456$ mm longitudinal length, and $0.76$ mm thickness of each ply. The material properties for the IM7/8552 AS4 considered in this application are reported in the data sheet published from the national center for advanced materials \cite{hexcel2016hexply}. To reproduce an operational condition, we consider a load applied along the major dimension of the plate which represents a simplified load condition of a wing panel during the flight. The damage consists in a cut of the fibers along the transversal direction in the third layer, and is selected to simulate a critical condition where the fault involves the layer with $0^{\circ}$ orientation that mainly contributes to support the load.

The health status of the system is represented through different damage parameters $\DesignVar = \left[ \FaultParam_1, \FaultParam_2, \FaultParam_3, \FaultParam_4  \right]$, including the transversal $\FaultParam_1$ and longitudinal $\FaultParam_2$ position of the cut, the extension of the cut $\FaultParam_3$ along the transversal direction, and the load $\FaultParam_4$ acting on the structure. Accordingly, the health monitoring task aims at identify the health status of the composite plate minimizing the discrepancy $\discrep$ between a real-world signal measured from the real system and the same signal computed evaluating a structural numerical model. For this procedure, we adopt the strain field $\Strain$ as the output signal to determine the health status of the structure: this signal is sensitive to failures in the fibers and can be easily measured in real-world applications and in laboratory. Formally, this health monitoring task is formulated as an inverse optimization problem: 

\begin{equation} \label{e: healthmonprob1}
    \MinDesignVar = \min_{\DesignVar \in \DesignSpace} \discrep (\DesignVar)
\end{equation}

\noindent where the discrepancy function $\discrep (\DesignVar)$ is computed as follows: 

 \begin{equation}
     \discrep (\DesignVar) = RMSE\left(\Strain_{ref}(\DesignVar), \Strain^{(\LevFid)}_{mon}(\DesignVar) \right) = \sqrt{\frac{1}{N} \sum_{j=1}^{N} \frac{(\Strain_{ref}^{j}(\DesignVar) - \Strain^{(\LevFid) j}_{mon}(\DesignVar))^2}{\Strain_{ref}^{j}(\DesignVar)}}
 \end{equation}
 
\noindent where $\Strain_{ref}(\DesignVar)$ is the reference strain field measured from the real system, $ \Strain^{(\LevFid)}_{mon}(\DesignVar)$ is the strain field computed with the $\LevFid$-th level of fidelity numerical model, and $N$ is the number of elements of the $\LevFid$-th level of fidelity field. The domain of the fault parameters $\DesignSpace = \Interval_{\FaultParam_1} \times \Interval_{\FaultParam_2} \times \Interval_{\FaultParam_3} \times \Interval_{\FaultParam_4}$ bounds the transversal $\Interval_{\FaultParam_1} = [0, 102]$ mm and longitudinal $\Interval_{\FaultParam_2} = [0, 456]$ mm position of the cut according to the maximum dimensions of the plate, while the intervals for the length of the cut $\Interval_{\FaultParam_3} = [0, 30]$ mm and the load $\Interval_{\FaultParam_4} = [0, 20]$ N are imposed from the expert knowledge about the specific structural health monitoring problem.

\subsection{Structural Models}
\label{s: structmodel}

The strain field $\Strain$ of the composite plate is modeled through the Reissner-Mindlin plate equations \cite{mindlin1951influence} and numerically solved adopting the Finite Element Method (FEM). The structural modeling approach represents the composite material of the undamaged structure as an orthotropic material assigning the properties of the carbon prepreg IM7/8552 AS4, and the cut in the fiber is modeled as an homogeneous material with the mechanical properties of the matrix. The boundary conditions impose a clamp in the lower section and a displacement in the upper portion of the plate with a region extended for the $10 \%$ of the total longitudinal length. This represents a simplification of the aerodynamic load acting on a composite panel adopted for the skin of an aircraft wing.

We use the software MSC Patran and MSC Nastran to develop two FEM models and compute the strain field of the damaged composite plate at different levels of fidelity. The high-fidelity model consists of a three-dimensional representation of the structure discretized through an adaptive grid of HEXA8 3D elements characterized by a dimension of $1$ mm in both the longitudinal and transversal direction near the cut region thought the border, and an increasingly coarse discretization away from the cut. This permits to capture the variation of the strain field that occurs in a small region near the cut with an high level of accuracy, while containing the overall computational cost reducing the number of elements far from the damaged location. The thickness of the plate is modeled inserting three HEXA8 elements for each of the four layers along the thickness direction to further enhance the accurate representation of the strain field. The cut is represented as a rectangular parallelepiped in the third layer characterized by a transversal extension discretized with HEXA8 elements, longitudinal extension equal to one element and thickness of three elements. The low-fidelity model approximates the composite plate through a two-dimensional representation discretized using an adaptive mesh of QUAD4 elements with transversal dimension of $2$ mm and longitudinal dimension of $4$ mm around the cut, and progressively increases the coarseness towards the boundaries of the plate. The cut is modeled through the same methodology of the accurate numerical model.

The high-fidelity model provides a reliable representation of the strain field as a result of the refined computational grid near the damage. This guarantees an high sensitivity to small incipient faults for which the variation of the strain field occurs in a contained region around the cut. In addition, this model allows to distinguish variations in the strain field caused by the application of intense loads in presence of a small cut in fiber -- which leads to a significant variation of the strain field even in regions far from the damage -- from an extended cut of the fiber -- which produces large strains in an extended region due to the size of the damage. We consider the high-fidelity structural model as an emulator of the real-world composite plate that is adopted to compute the reference strain field $\Strain_{ref}(\DesignVar)$, and is used as the highest level of fidelity available to evaluate the monitoring signal $\Strain_{mon}^{(\MaxLevFid=2)}(\DesignVar)$. The low-fidelity representation reduces the computational burden if compared with the high-fidelity model, and achieves a satisfactory accuracy of the strain field prediction for damages characterized by an extended cut in the fiber. However, the coarse discretization entails an inaccurate evaluation of the strain field for small incipient damages of the composite plate, and fails in differentiating the increase of the strain associated with small cuts in presence of significant loads from extended damages in the fiber. This results in an approximated representation of the monitoring strain signal $\Strain_{mon}^{(\LevFid=1)}(\DesignVar)$. 

For this set of structural models, the relative computational cost $\CompCost^{(\LevFid)}$ is measured as the average CPU time required to solve the structural model at $\LevFid$-th level of fidelity referred to the high-fidelity CPU time. From our preliminary experiments, we observe that the high-fidelity model requires $40$ minutes to achieve convergence while the low-fidelity representation takes $8$ minutes on a single core of a desktop PC with Intel Core i7-8700 (3.2 GHz) and 32 GB of RAM. Accordingly, we set $\CompCost^{(\MaxLevFid)}=1$ for the high-fidelity structural model and $\CompCost^{(1)}=0.2$ for the low-fidelity structural model. 

\begin{figure*}[t!]
    \centering
     \subfigure[]{%
        \includegraphics[width=0.29\linewidth,trim=15cm 0 15cm 0, clip]{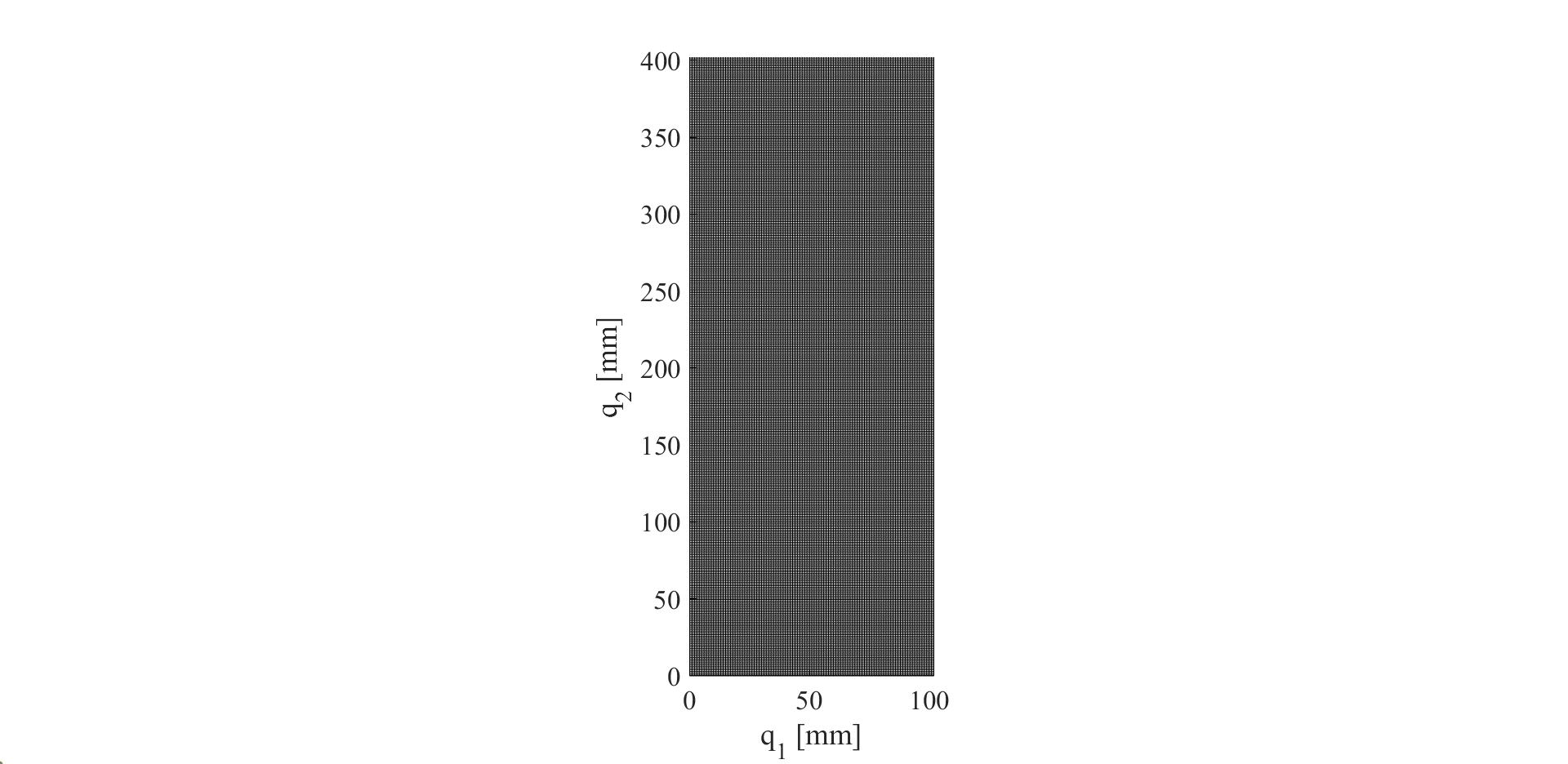} 
        \label{fig:hfmesh}}
        \subfigure[]{%
        \includegraphics[width=0.65\linewidth,trim=3cm 0 1cm 0, clip]{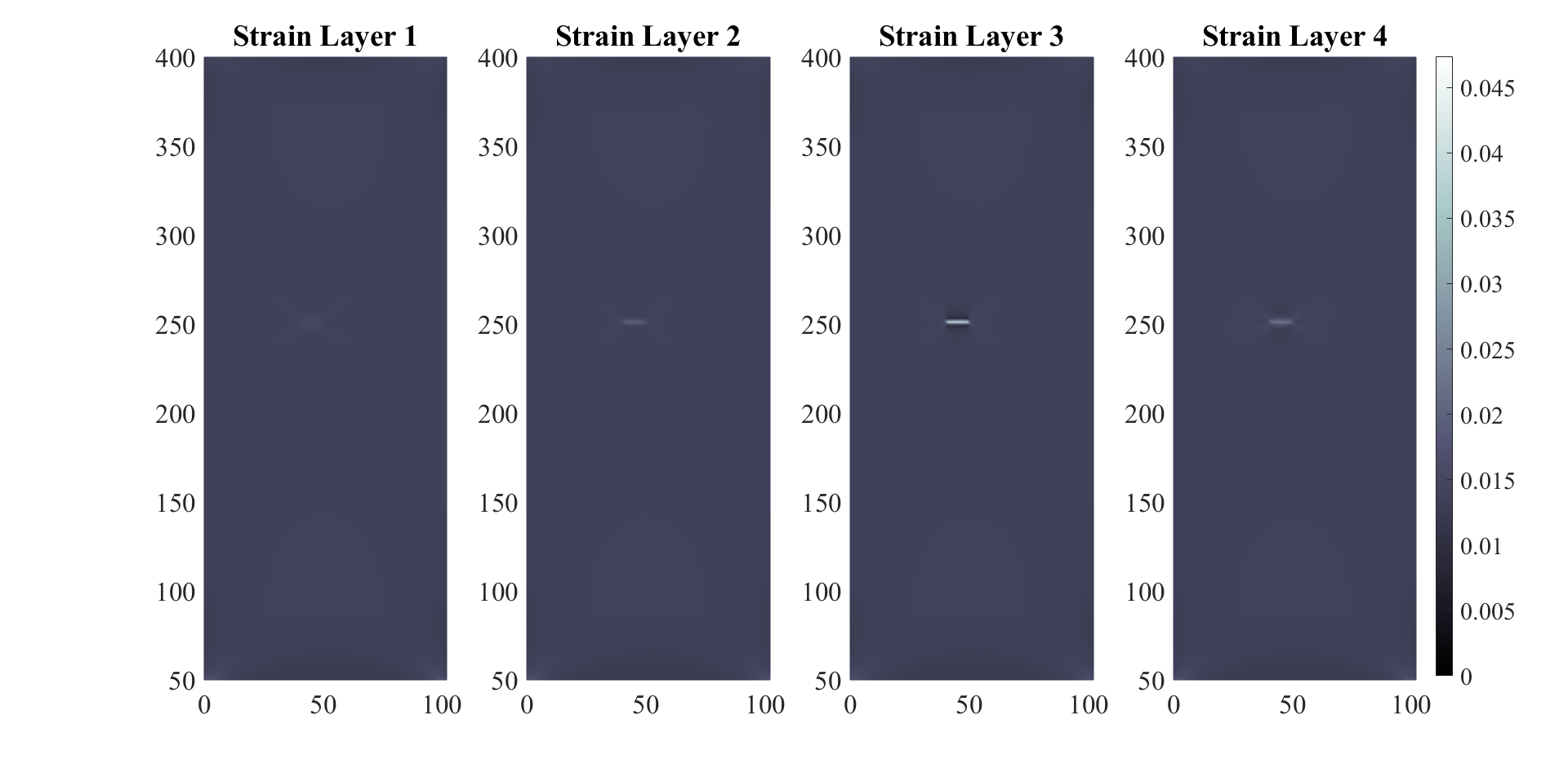}
        \label{fig:hfstruct}}

    \caption{(a) high-fidelity discretization of the computational domain, and (b) high-fidelity strain distribution for the four layers of the damaged composite plate.}\label{fig:hfmodel}
\end{figure*}

\begin{figure*}[t!]
    \centering
     \subfigure[]{%
        \includegraphics[width=0.29\linewidth,trim=15cm 0 15cm 0, clip]{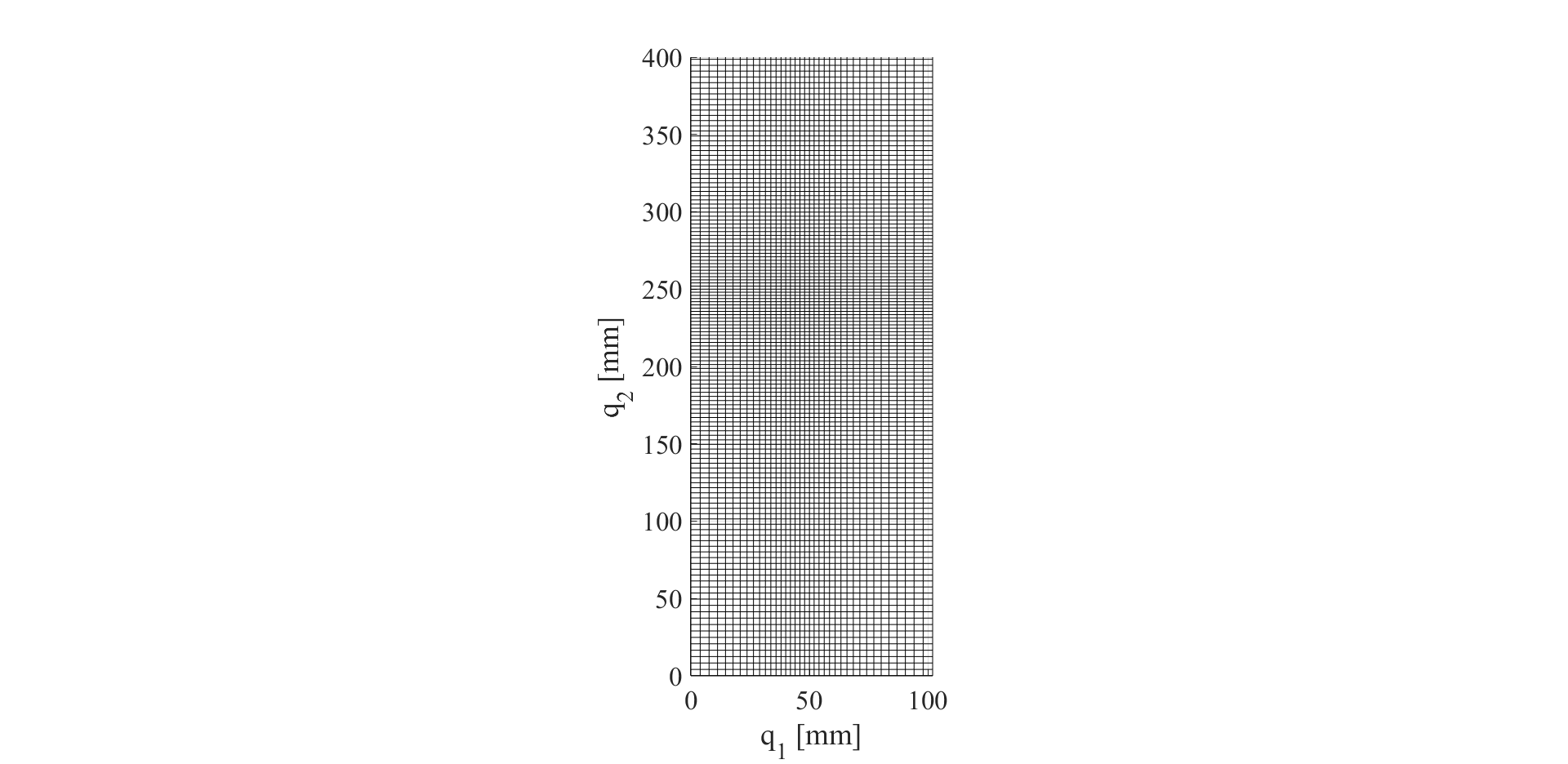} 
        \label{fig:lfmesh}}
        \subfigure[]{%
        \includegraphics[width=0.65\linewidth,trim=3cm 0 1cm 0, clip]{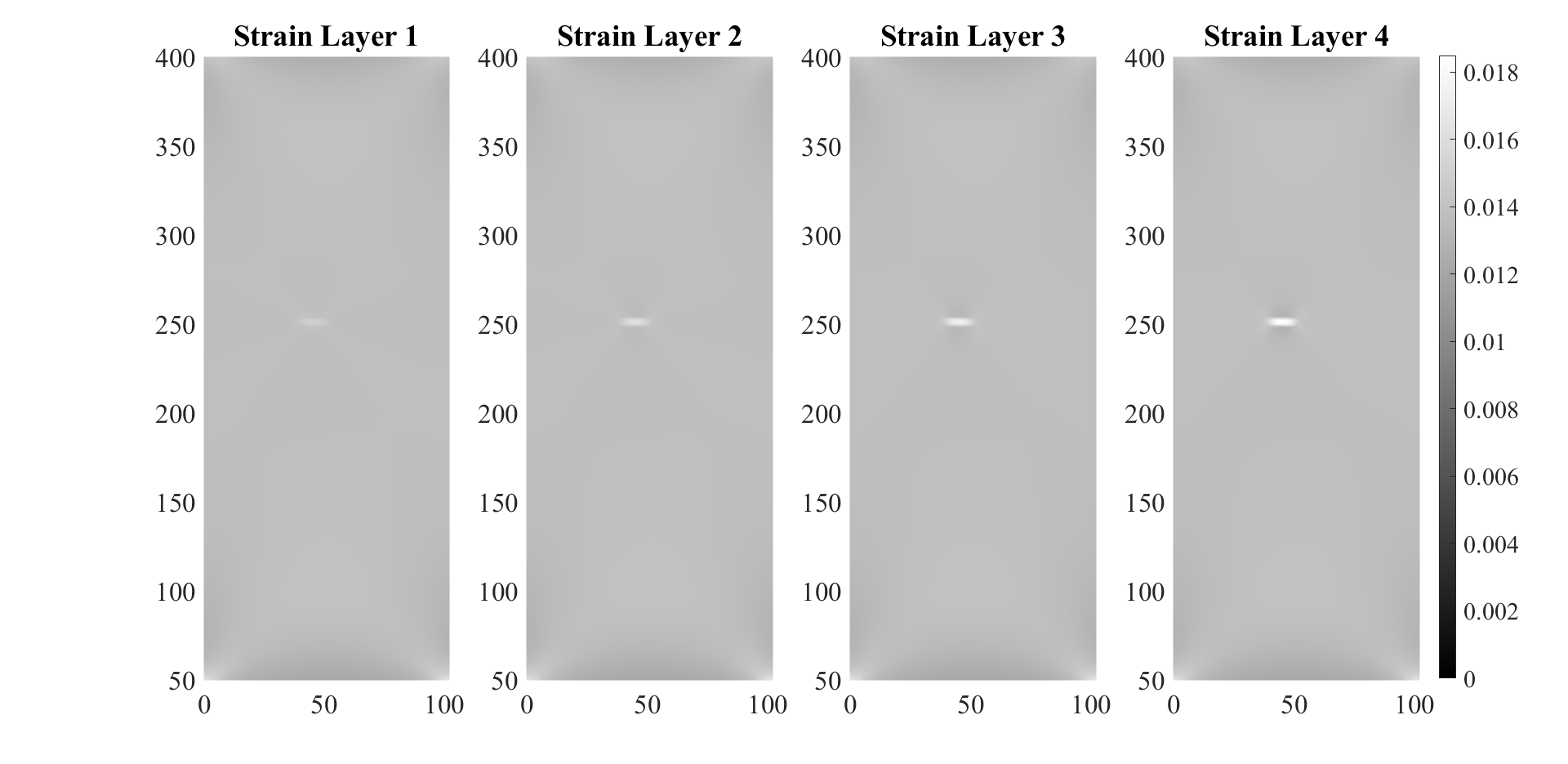}
        \label{fig:lfstruct}}

    \caption{(a) low-fidelity discretization of the computational domain, and (b) low-fidelity strain distribution for the four layers of the damaged composite plate.}\label{fig:lfmodel}
\end{figure*}

Figure \ref{fig:hfmodel} and Figure \ref{fig:lfmodel} illustrate the computational mesh and strain field over the four layers of the plate computed with the high-fidelity and low-fidelity structural model, respectively. These results are achieved for a cut in the fiber of the third layer located horizontally at 40 mm and vertically at 250 mm considering a cut length of 10 mm and load equal to 5 N.

\subsection{Structural Physics-Aware Utility Function}
\label{s:structutility}

The physics-aware utility function is conceived to incorporate expert knowledge about the appropriate structural model to be evaluated in presence of a small incipient cut concurrently with a significant load condition, or an extended damage in the fiber of the composite plate. This is realized through a bias in the search procedure $\PhysicsVec = \left[ \FaultParam_3, \FaultParam_4 \right]$ that encodes the specific structure of the domain, and is induced by the length of the fiber cut $\FaultParam_3$ and the load applied on the plate $\FaultParam_4$. Thus, we formalize $\alpha_4(\FaultParam_3, \FaultParam_4, \LevFid)$ for the health monitoring problem as follows:

\begin{equation} \label{e:MFAF7}
\begin{split}
\alpha_4 (\FaultParam_3, \FaultParam_4, \LevFid) =
 & \left\{ \begin{array}{lll}  1 & \quad \mbox{if} & \LevFid = 1,...,\MaxLevFid-1  \\  0.5\frac{\FaultParam_{3max}}{\FaultParam_3} + 0.5\frac{1}{\FaultParam_{4max} - \FaultParam_4} & \quad \mbox{if} & \LevFid = \MaxLevFid   \end{array} \right.
\end{split}
\end{equation}

\noindent where $\FaultParam_{3max}$ is the maximum length of the cut in the fiber and $\FaultParam_{4max}$ is the maximum load applied on the plate. This physics-aware utility function realizes a sort of expert reasoning and privileges the evaluation of the high-fidelity structural model for small incipient damages and high load values. Indeed, $\alpha_4 (\FaultParam_3, \FaultParam_4, \LevFid)$ increases the value of the multifidelity acquisition function (Equation \eqref{e:MFAF1}) when the health status of the plate affected by an incipient cut and high load condition is evaluated with the high-fidelity numerical model. This ensures an accurate estimate of the strain field and permits to distinguish a narrow cut that generates large variations of the strain field amplified by high loading conditions form the magnification of strains generated by an extended cut in the fiber.

\subsection{Structural Health Monitoring Results}

This section reports and discusses the results achieved with the PA-MFBO framework for the structural health monitoring example to evaluate the capabilities of the algorithm against a damage identification problem. The outcomes of the PA-MFBO are compared against the efficient global optimization (EGO) algorithm \cite{jones1998efficient} and the multifidelity Bayesian optimization implementing the multifidelity expected improvement (MFBO) \cite{huang2006sequential}. To assess the performance of the optimization algorithms on this test case, we compute the following assessment metrics:

\begin{equation}
    \Error(\FaultParam_i) = \frac{\vert \FaultParam_i^*-\FaultParam_i \vert}{\FaultParam_i^*} \cdot 100
\end{equation}

\begin{equation}
     \discrep^{*} = \min (\discrep(\DesignVar))
\end{equation}

\noindent where $\FaultParam_i^*$ is the actual level of damage that affects the composite plate, $\FaultParam_i$ is the level of damage inferred by the algorithm considering the $i$-th fault parameter, and $\discrep(\DesignVar)$ is the value of the discrepancy between the reference strain signal and the strain field computed with the high-fidelity model. The percentage relative error $\Error(\FaultParam_i)$ quantifies the accuracy related to the identification of the faults parameters, and $\discrep^{*}$ represents the minimum value of the discrepancy computed by the algorithms and provides a measure of the improvement in the solution of the optimization procedure.

\begin{figure*}[t!]
    \centering

    \subfigure[]{%
        \includegraphics[width=0.35\linewidth,trim=220 0 245 0, clip]{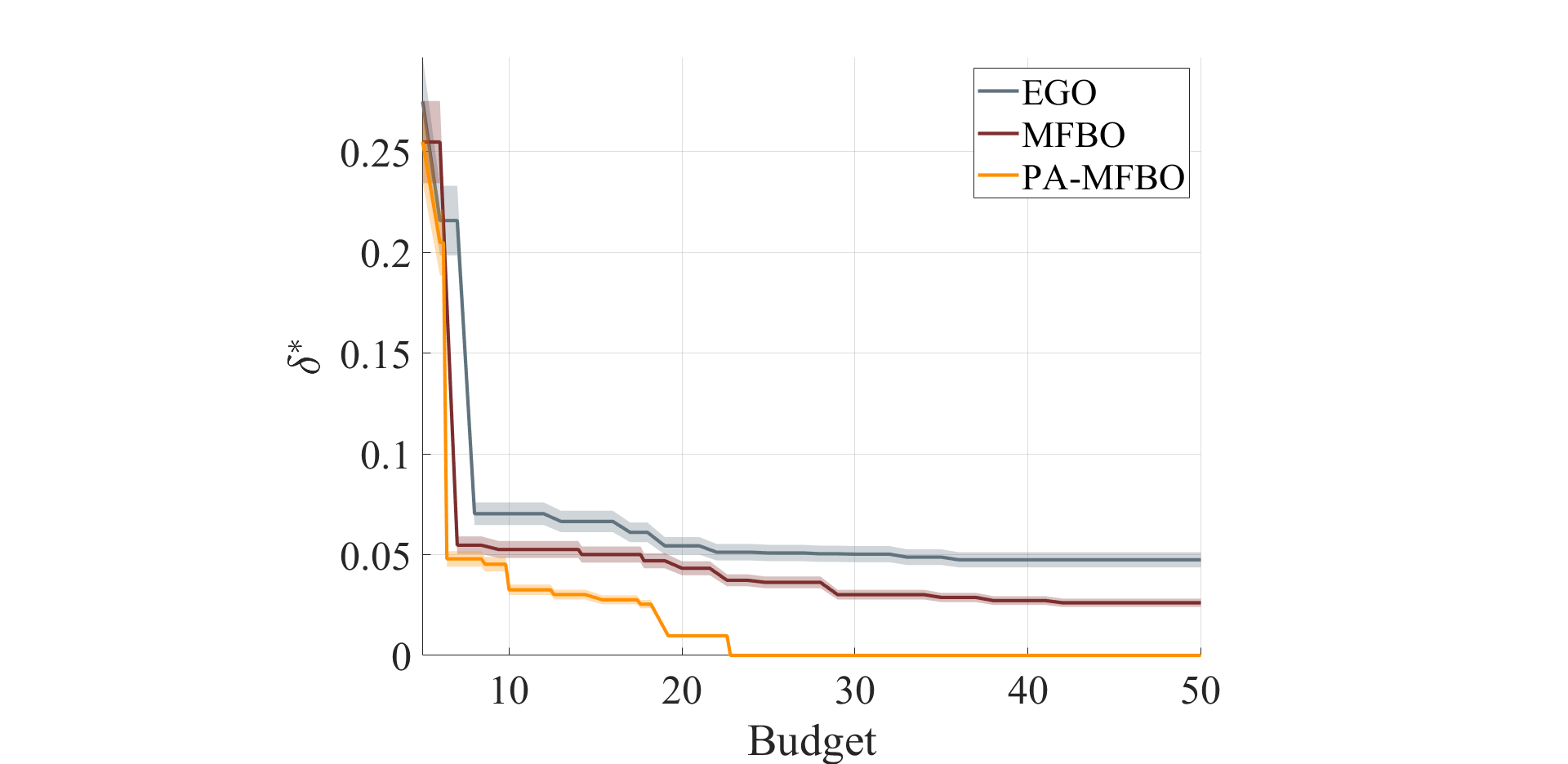} 
        \label{fig:Dis}}

    \subfigure[]{%
        \includegraphics[width=0.35\linewidth,trim=220 0 245 0, clip]{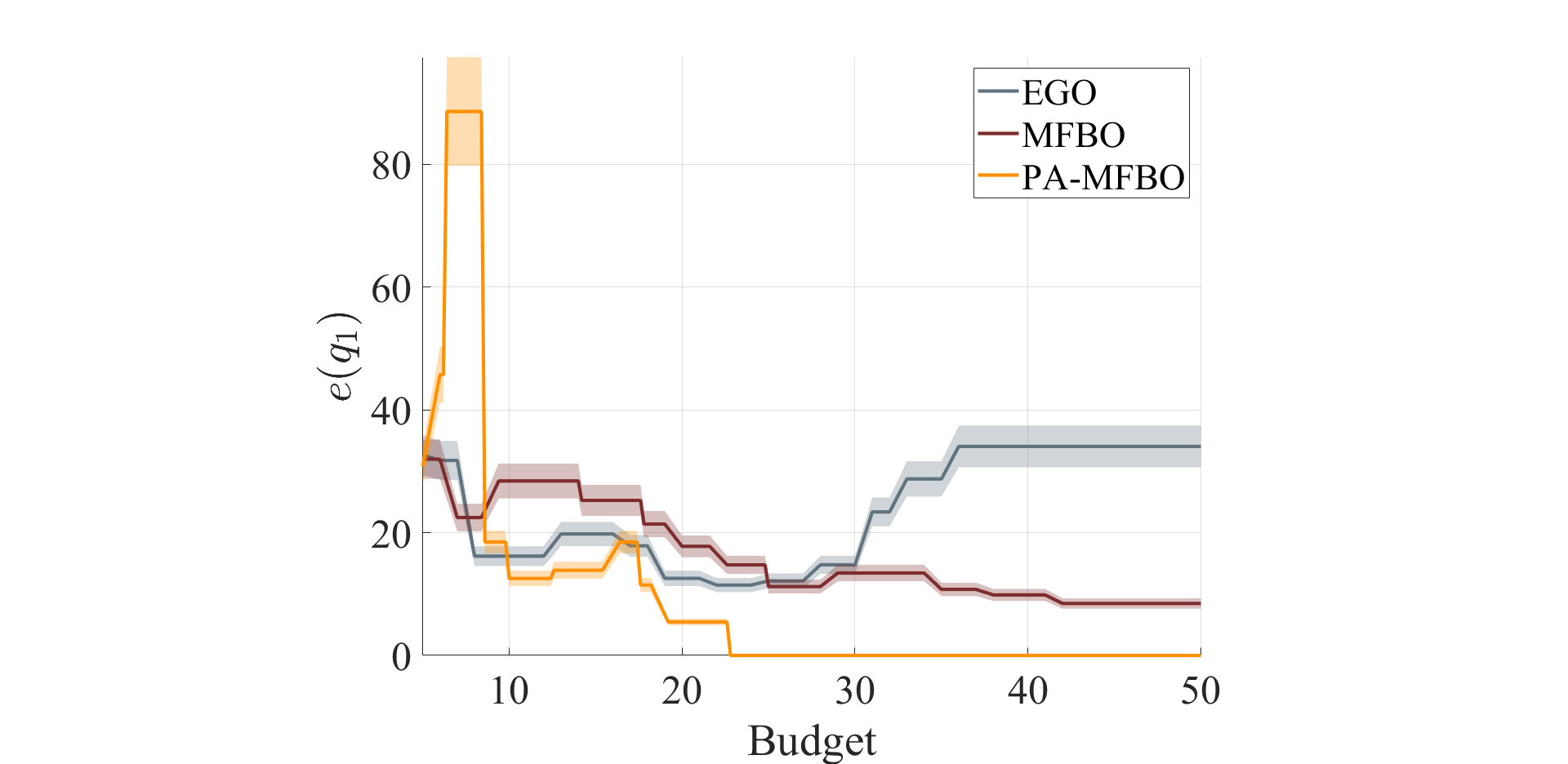}
        \label{fig:k1}}
     \subfigure[]{%
        \includegraphics[width=0.35\linewidth,trim=220 0 245 0, clip]{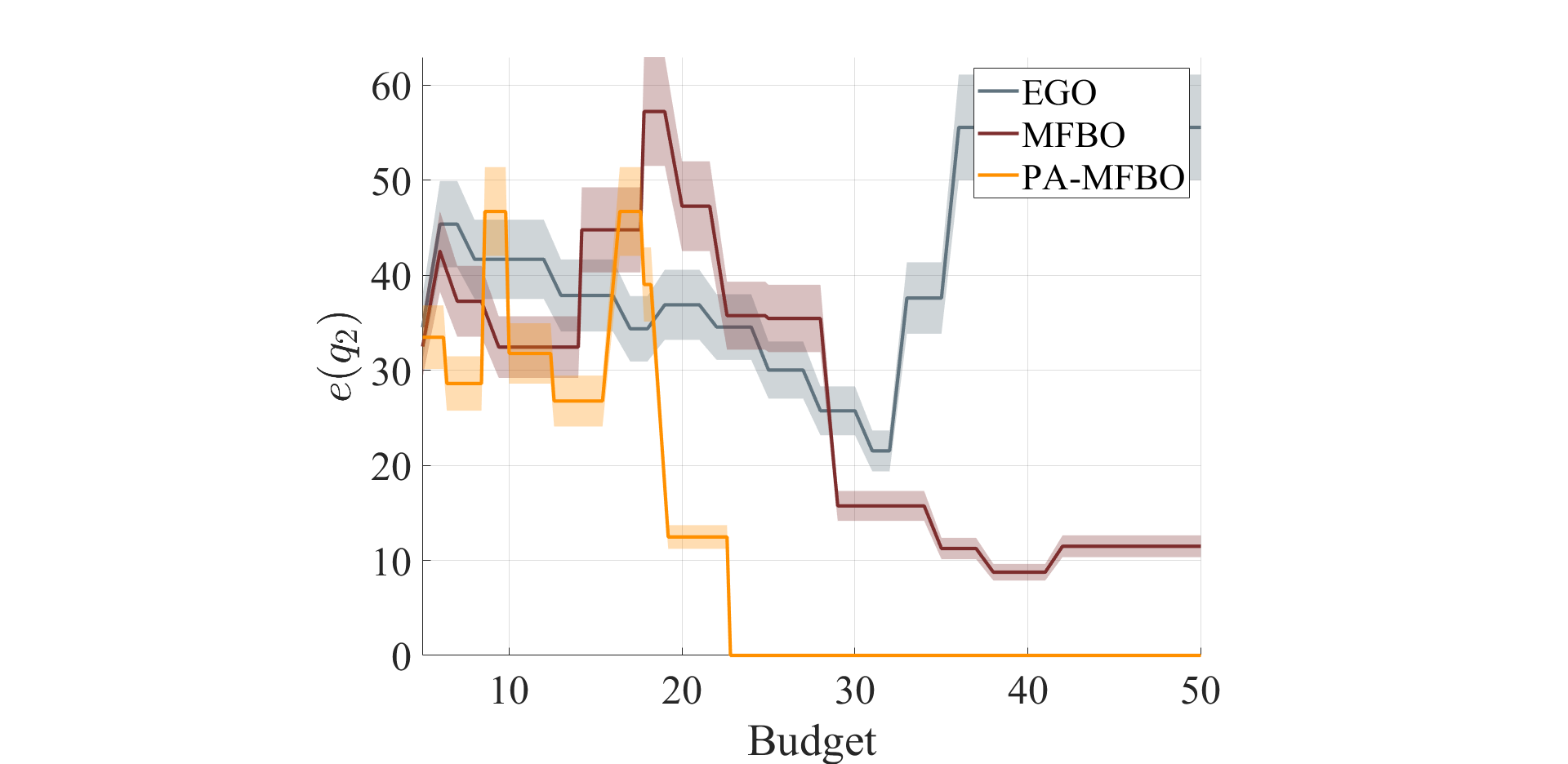}
        \label{fig:k2}}
        
    \subfigure[]{%
        \includegraphics[width=0.35\linewidth,trim=220 0 245 0, clip]{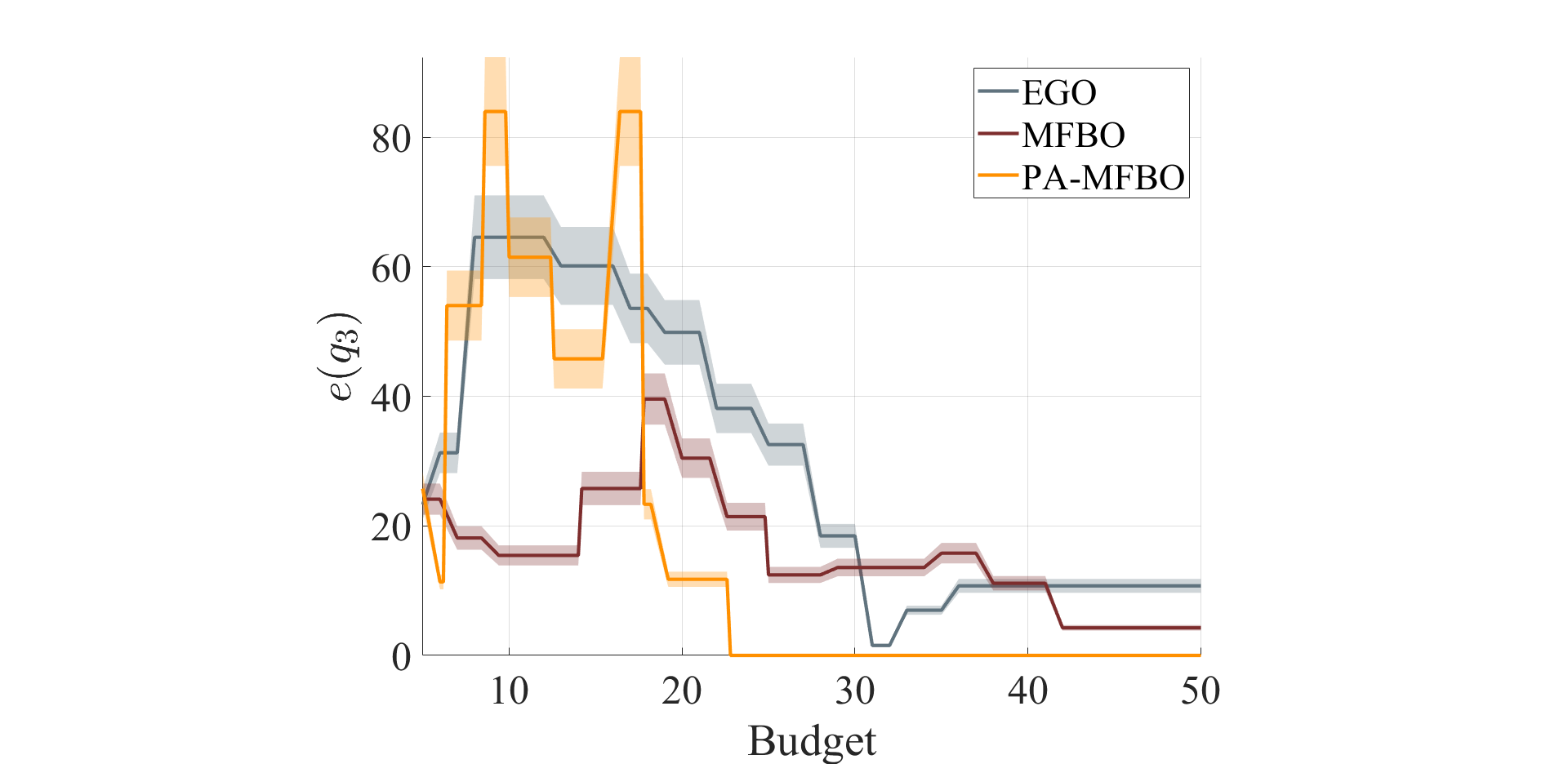}
        \label{fig:k3}}
    \subfigure[]{%
       \includegraphics[width=0.36\linewidth,trim=220 0 245 0, clip]{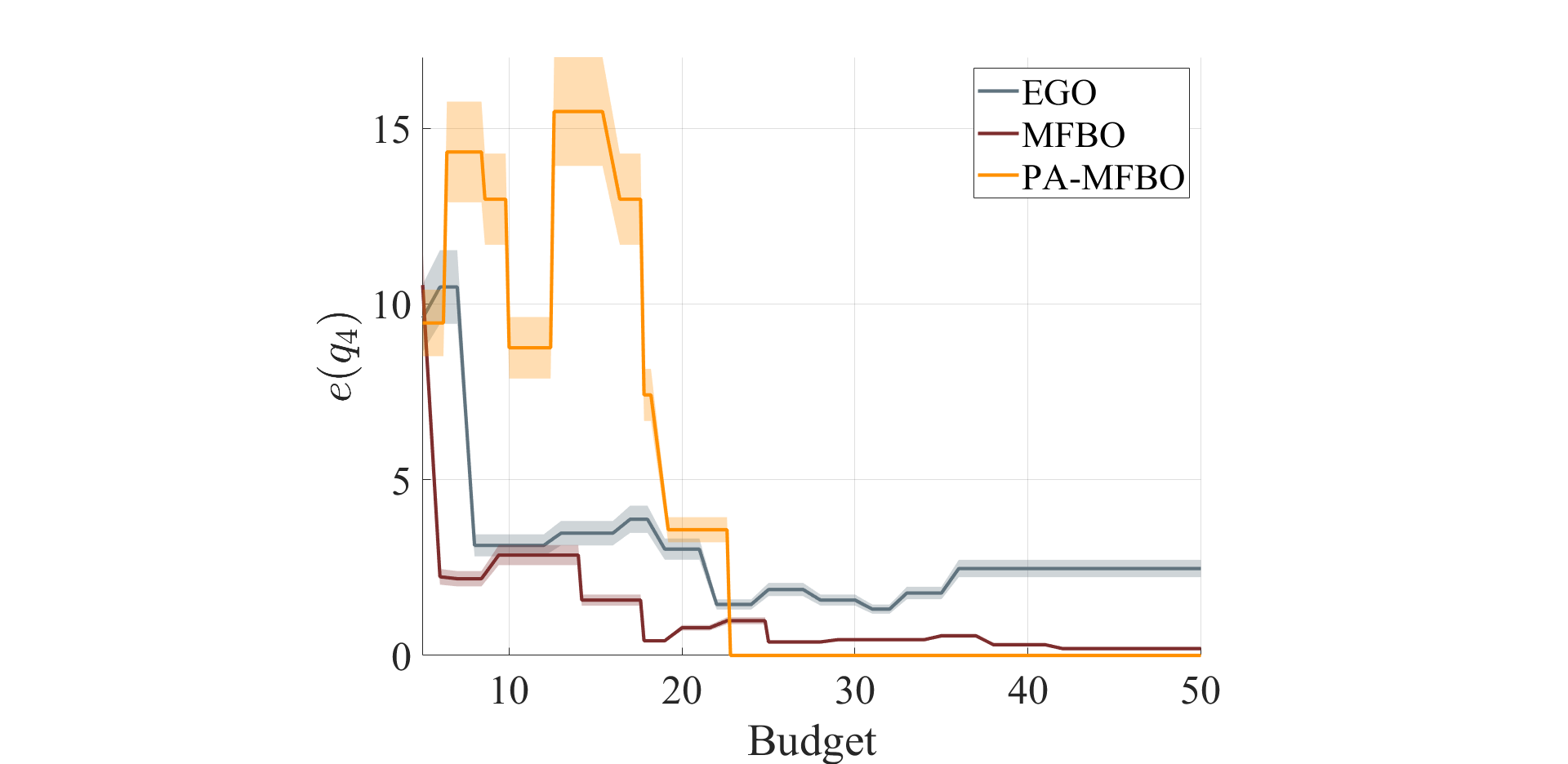}
        \label{fig:k4}}

    \caption{Statistics over 25 runs of the minimum discrepancy $\discrep^{*}$ and percentage relative error of the inference of the damage parameters $\Error(\FaultParam_i)$ obtained with the competing algorithms.}\label{fig:StructRes}
\end{figure*}

We consider a statistics over 25 different combinations of fault parameters determined through the scaled Latin hypercube sampling process proposed by Berri et al. \cite{berri2019real}. This design of experiments permits to increase the distribution of the fiber cut length located in proximity of the undamaged condition, and improves the amount of incipient damages evaluated during the experiments. In particular, the multifidelity methods start the health monitoring procedure with 17 damage configurations among which 15 faults are evaluated with the low-fidelity structural model and 2 with the high-fidelity representation, while the single-fidelity algorithm is initialized with 5 damages evaluated with the high-fidelity model.

Figure \ref{fig:StructRes} reports the outcomes in terms of median and interval between the 25-th and 75-th percentiles for both the assessment metrics, and Table \ref{t:StructConv} summarizes the convergence results both as functions of the computational budget $\Budget = \sum \CompCost^{(\LevFid)}_{\IterOpt}$ measured as the cumulative computational cost $\CompCost^{(\LevFid)}_{\IterOpt}$ used at each iteration $\IterOpt$ to evaluate the $\LevFid$-th structural level of fidelity. Overall, the multifidelity algorithms -- PA-MFBO and MFBO -- achieve lower values of the identification error rather than the single fidelity strategy implementing the high-fidelity structural model -- EGO. However, it can be noticed that the proposed PA-MFBO is the only optimization method capable to infer the exact health status of the composite plate ($\Error(\FaultParam_i) = 0 \%$) with a computational budget of just $\Budget=22.8$, which corresponds to less than half of the budget consumed by EGO and MFBO ($\Budget=50$) to converge to suboptimal values of the identification error $\Error(\FaultParam_i)>0 \%$. Table \ref{t:Struct_levfidcalls} illustrates the average number of aerodynamic models queries at different levels of fidelity for EGO, MFBO and PA-MFBO over the performed 25 tests. We notice that the multifidelity algorithms MFBO and PA-MFBO massively query the low-fidelity structural model to accelerate the exploration of different damage configurations and contain the evaluation of expensive high-fidelity analysis. However, PA-MFBO effectively evaluates the high-fidelity model to progressively reduce the inference error of damages, and accurately identify the health status of the composite plate adopting almost the same number of high-fidelity evaluation as MFBO which leads to non-negligible errors. These results suggest that the introduction of prior expert knowledge about the health monitoring problem enhances the accuracy of the damage identification procedure. A remarkable outcome is that the PA-MFBO algorithm is the only optimization framework capable to accurately identify the health status of the composite plate within the allocated budget. This outcome suggests that the inclusion of the expert knowledge about the structures of the domain and the behaviour of the numerical models over those structures allow to obtain a fast and robust inference performance.

\begin{table} [t!]
\centering
\begin{tabular}{lccccc}
\hline\noalign{\smallskip}
Method & $\Error(\FaultParam_1)$ & $\Error(\FaultParam_2)$ & $\Error(\FaultParam_3)$ & $\Error(\FaultParam_4)$ & $\discrep^{*}$ \\
\noalign{\smallskip}\hline\noalign{\smallskip}
PA-MFBO      & $0.00\%$ &  $0.00\%$  & $0.00\%$  & $0.00\%$ & $0.00$  \\
MFBO         & $8.46\%$ &  $11.5\%$  & $4.26\%$  & $0.19\%$  & $0.0261$  \\
EGO          & $34.1\%$ &  $55.5\%$  & $10.7\%$  & $2.47\%$  & $0.0474$  \\
\noalign{\smallskip}\hline
\end{tabular}
\caption{Convergence results of the percentage relative error of the inference of the damage parameters $\Error(\FaultParam_i)$, and minimum discrepancy value $\discrep^*$ obtained with the competing algorithms.}
\label{t:StructConv} 
\end{table}

\begin{table} [t!]
\centering     
\begin{tabular}{lcc}
\hline\noalign{\smallskip}
Method & HF evaluations & LF evaluations  \\
\noalign{\smallskip}\hline\noalign{\smallskip} 
EGO      & 50 & - \\
MFBO     &   7.7  & 211.5  \\
PA-MFBO  &  8.1   &  209.5 \\
\noalign{\smallskip}\hline
\end{tabular}
\caption{Average evaluations of the high-fidelity (HF) and low-fidelity (LF) structural model over the 25 runs of the structural health monitoring problem.}
\label{t:Struct_levfidcalls}
\end{table}

\section{Concluding Remarks}
\label{s:Conclusions}

This paper recognizes that domain knowledge is commonly available in science and engineering, and can be used to accelerate and improve the multifidelity optimization process. We propose a Physics-Aware Multifidelity Bayesian Optimization -- PA-MFBO -- framework that incorporates forms of prior scientific and expert knowledge about the physical domain during the search procedure. This is achieved introducing a learning bias formalized as a physics-aware multifidelity acquisition function that leverages the knowledge about the structure of the domain to enhance the accuracy of the solution and alleviate the computational cost for optimization.

The results achieved with the PA-MFBO are observed and discussed for an aerodynamic design problem and a structural health monitoring problem. In the design test-case, the PA-MFBO introduces a bias to pursue the awareness about the transition of fluid regimes through the Mach number. In the health assessment task, PA-MFBO incorporates a bias based on the expert knowledge about the features of the domain for specific combinations of load and extension of the damage. We note that for the two optimization problems our methodology outperforms standard single-fidelity and multifidelity Bayesian algorithms in terms of accuracy and acceleration of the search. In particular, the PA-MFBO identifies aerodynamic design solutions capable to deliver a performance improvement of the $24.31 \%$ in less than half the computational time required by competing algorithms to search suboptimal designs. Moreover, PA-MFBO is the only algorithm that permits the robust identification of damages, which otherwise would have required more computational resources or might have led to an inaccurate health assessment. 

Overall, the results show the importance of embedding forms of prior knowledge in design and health monitoring optimization procedures as an enabling technique to satisfy the ever-increasing demand for performance and reliability to meet sustainability goals.

\section*{Acknowledgments}
\label{s:Acknowledgments}

This work was supported by project Multisource Frameworks to Support Real-time Structural Assessment and Autonomous Decision Making under the Visiting Professor program of Politecnico di Torino, and by the University’s Doctoral Scholarship. Additional acknowledgement to Prof. Paolo Maggiore and the LABINF HPC Center of Politecnico di Torino for the support and the access to the computing resources. The authors thank Mr. Mirko Ermacora and Mr. Matteo Fenoglio for their assistance with the structural models of the composite plate.

\bibliographystyle{unsrt}  
\bibliography{references}

\begin{thebibliography}{10}

\bibitem{martins2021engineering}
Joaquim~RRA Martins and Andrew Ning.
\newblock {\em Engineering design optimization}.
\newblock Cambridge University Press, 2021.

\bibitem{forrester2007multi}
Alexander~IJ Forrester, Andr{\'a}s S{\'o}bester, and Andy~J Keane.
\newblock Multi-fidelity optimization via surrogate modelling.
\newblock {\em Proceedings of the royal society a: mathematical, physical and engineering sciences}, 463(2088):3251--3269, 2007.

\bibitem{park2017remarks}
Chanyoung Park, Raphael~T Haftka, and Nam~H Kim.
\newblock Remarks on multi-fidelity surrogates.
\newblock {\em Structural and Multidisciplinary Optimization}, 55:1029--1050, 2017.

\bibitem{peherstorfer2018survey}
Benjamin Peherstorfer, Karen Willcox, and Max Gunzburger.
\newblock Survey of multifidelity methods in uncertainty propagation, inference, and optimization.
\newblock {\em Siam Review}, 60(3):550--591, 2018.

\bibitem{beran2020comparison}
Philip~S Beran, Dean Bryson, Andrew~S Thelen, Matteo Diez, and Andrea Serani.
\newblock Comparison of multi-fidelity approaches for military vehicle design.
\newblock In {\em AIAA AVIATION 2020 FORUM}, page 3158, 2020.

\bibitem{thelen2022multi}
Andrew~S Thelen, Dean~E Bryson, Bret~K Stanford, and Philip~S Beran.
\newblock Multi-fidelity gradient-based optimization for high-dimensional aeroelastic configurations.
\newblock {\em Algorithms}, 15(4):131, 2022.

\bibitem{tezzele2023multifidelity}
Marco Tezzele, Lorenzo Fabris, Matteo Sidari, Mauro Sicchiero, and Gianluigi Rozza.
\newblock A multifidelity approach coupling parameter space reduction and nonintrusive pod with application to structural optimization of passenger ship hulls.
\newblock {\em International Journal for Numerical Methods in Engineering}, 124(5):1193--1210, 2023.

\bibitem{anselma2020multidisciplinary}
PG~Anselma, C~Boursier Niutta, L~Mainini, and G~Belingardi.
\newblock Multidisciplinary design optimization for hybrid electric vehicles: component sizing and multi-fidelity frontal crashworthiness.
\newblock {\em Structural and Multidisciplinary Optimization}, 62:2149--2166, 2020.

\bibitem{lai2022building}
Xiaonan Lai, Xiwang He, Shuo Wang, Xiaobang Wang, Wei Sun, and Xueguan Song.
\newblock Building a lightweight digital twin of a crane boom for structural safety monitoring based on a multifidelity surrogate model.
\newblock {\em Journal of Mechanical Design}, 144(6):064502, 2022.

\bibitem{makkar2022machine}
Gaurav Makkar, Cameron Smith, George Drakoulas, Fotis Kopsaftopoulos, and Farhan Gandhi.
\newblock A machine learning framework for physics-based multi-fidelity modeling and health monitoring for a composite wing.
\newblock In {\em ASME International Mechanical Engineering Congress and Exposition}, volume 86625, page V001T01A008. American Society of Mechanical Engineers, 2022.

\bibitem{viana2014special}
Felipe~AC Viana, Timothy~W Simpson, Vladimir Balabanov, and Vasilli Toropov.
\newblock Special section on multidisciplinary design optimization: metamodeling in multidisciplinary design optimization: how far have we really come?
\newblock {\em AIAA journal}, 52(4):670--690, 2014.

\bibitem{guo2018analysis}
Zhendong Guo, Liming Song, Chanyoung Park, Jun Li, and Raphael~T Haftka.
\newblock Analysis of dataset selection for multi-fidelity surrogates for a turbine problem.
\newblock {\em Structural and Multidisciplinary Optimization}, 57:2127--2142, 2018.

\bibitem{song2019general}
Jialin Song, Yuxin Chen, and Yisong Yue.
\newblock A general framework for multi-fidelity bayesian optimization with gaussian processes.
\newblock In {\em The 22nd International Conference on Artificial Intelligence and Statistics}, pages 3158--3167. PMLR, 2019.

\bibitem{movckus1975bayesian}
Jonas Mo{\v{c}}kus.
\newblock On bayesian methods for seeking the extremum.
\newblock In {\em Optimization Techniques IFIP Technical Conference: Novosibirsk, July 1--7, 1974}, pages 400--404. Springer, 1975.

\bibitem{snoek2012practical}
Jasper Snoek, Hugo Larochelle, and Ryan~P Adams.
\newblock Practical bayesian optimization of machine learning algorithms.
\newblock {\em Advances in neural information processing systems}, 25, 2012.

\bibitem{frazier2018bayesian}
Peter~I Frazier.
\newblock Bayesian optimization.
\newblock In {\em Recent advances in optimization and modeling of contemporary problems}, pages 255--278. Informs, 2018.

\bibitem{meliani2019multi}
Mostafa Meliani, Nathalie Bartoli, Thierry Lefebvre, Mohamed-Amine Bouhlel, Joaquim~RRA Martins, and Joseph Morlier.
\newblock Multi-fidelity efficient global optimization: Methodology and application to airfoil shape design.
\newblock In {\em AIAA aviation 2019 forum}, page 3236, 2019.

\bibitem{tran2020multi}
Anh Tran, Julien Tranchida, Tim Wildey, and Aidan~P Thompson.
\newblock Multi-fidelity machine-learning with uncertainty quantification and bayesian optimization for materials design: Application to ternary random alloys.
\newblock {\em The Journal of Chemical Physics}, 153(7):074705, 2020.

\bibitem{serani2019adaptive}
A~Serani, R~Pellegrini, J~Wackers, C-E Jeanson, P~Queutey, Michel Visonneau, and M~Diez.
\newblock Adaptive multi-fidelity sampling for cfd-based optimisation via radial basis function metamodels.
\newblock {\em International Journal of Computational Fluid Dynamics}, 33(6-7):237--255, 2019.

\bibitem{perdikaris2016model}
Paris Perdikaris and George~Em Karniadakis.
\newblock Model inversion via multi-fidelity bayesian optimization: a new paradigm for parameter estimation in haemodynamics, and beyond.
\newblock {\em Journal of The Royal Society Interface}, 13(118):20151107, 2016.

\bibitem{di2021multifidelity}
Francesco Di~Fiore, Paolo Maggiore, and Laura Mainini.
\newblock Multifidelity domain-aware learning for the design of re-entry vehicles.
\newblock {\em Structural and Multidisciplinary Optimization}, 64(5):3017--3035, 2021.

\bibitem{di2022non}
Francesco Di~Fiore and Laura Mainini.
\newblock Nm-mf: Non-myopic multifidelity framework for constrained multi-regime aerodynamic optimization.
\newblock {\em AIAA Journal}, 61(3):1270--1280, 2023.

\bibitem{williams1995gaussian}
Christopher Williams and Carl Rasmussen.
\newblock Gaussian processes for regression.
\newblock {\em Advances in neural information processing systems}, 8, 1995.

\bibitem{schulz2018tutorial}
Eric Schulz, Maarten Speekenbrink, and Andreas Krause.
\newblock A tutorial on gaussian process regression: Modelling, exploring, and exploiting functions.
\newblock {\em Journal of Mathematical Psychology}, 85:1--16, 2018.

\bibitem{kennedy2000predicting}
Marc~C Kennedy and Anthony O'Hagan.
\newblock Predicting the output from a complex computer code when fast approximations are available.
\newblock {\em Biometrika}, 87(1):1--13, 2000.

\bibitem{jones1998efficient}
Donald~R Jones, Matthias Schonlau, and William~J Welch.
\newblock Efficient global optimization of expensive black-box functions.
\newblock {\em Journal of Global optimization}, 13(4):455, 1998.

\bibitem{kushner1964new}
Harold~J Kushner.
\newblock A new method of locating the maximum point of an arbitrary multipeak curve in the presence of noise.
\newblock {\em Journal of Basic Engineering}, 86(1):97--106, 1964.

\bibitem{hennig2012entropy}
Philipp Hennig and Christian~J Schuler.
\newblock Entropy search for information-efficient global optimization.
\newblock {\em Journal of Machine Learning Research}, 13(6), 2012.

\bibitem{huang2006sequential}
Deng Huang, Theodore~T Allen, William~I Notz, and R~Allen Miller.
\newblock Sequential kriging optimization using multiple-fidelity evaluations.
\newblock {\em Structural and Multidisciplinary Optimization}, 32(5):369--382, 2006.

\bibitem{ruan2020variable}
Xiongfeng Ruan, Ping Jiang, Qi~Zhou, Jiexiang Hu, and Leshi Shu.
\newblock Variable-fidelity probability of improvement method for efficient global optimization of expensive black-box problems.
\newblock {\em Structural and Multidisciplinary Optimization}, 62(6):3021--3052, 2020.

\bibitem{zhang2017information}
Yehong Zhang, Trong~Nghia Hoang, Bryan Kian~Hsiang Low, and Mohan Kankanhalli.
\newblock Information-based multi-fidelity bayesian optimization.
\newblock In {\em NIPS Workshop on Bayesian Optimization}, 2017.

\bibitem{takeno2020multi}
Shion Takeno, Hitoshi Fukuoka, Yuhki Tsukada, Toshiyuki Koyama, Motoki Shiga, Ichiro Takeuchi, and Masayuki Karasuyama.
\newblock Multi-fidelity bayesian optimization with max-value entropy search and its parallelization.
\newblock In {\em International Conference on Machine Learning}, pages 9334--9345. PMLR, 2020.

\bibitem{ForresterAl2008}
Alexander Forrester, Andras Sobester, and Andy Keane.
\newblock {\em Engineering design via surrogate modelling: a practical guide}.
\newblock John Wiley \& Sons, 2008.

\bibitem{drela1998pros}
Mark Drela.
\newblock Pros and cons of airfoil optimization.
\newblock {\em Frontiers of computational fluid dynamics}, 1998:363--381, 1998.

\bibitem{li2002robust}
Wu~Li, Luc Huyse, and Sharon Padula.
\newblock Robust airfoil optimization to achieve drag reduction over a range of mach numbers.
\newblock {\em Structural and Multidisciplinary Optimization}, 24(1):38--50, 2002.

\bibitem{elger2020engineering}
Donald~F Elger, Barbara~A LeBret, Clayton~T Crowe, and John~A Roberson.
\newblock {\em Engineering fluid mechanics}.
\newblock John Wiley \& Sons, 2020.

\bibitem{young2010brief}
Donald~F Young, Bruce~R Munson, Theodore~H Okiishi, and Wade~W Huebsch.
\newblock {\em A brief introduction to fluid mechanics}.
\newblock John Wiley \& Sons, 2010.

\bibitem{quagliarella2020open}
Domenico Quagliarella and Matteo Diez.
\newblock An open-source aerodynamic framework for benchmarking multi-fidelity methods.
\newblock In {\em AIAA AVIATION 2020 FORUM}, page 3179, 2020.

\bibitem{economon2016su2}
Thomas~D Economon, Francisco Palacios, Sean~R Copeland, Trent~W Lukaczyk, and Juan~J Alonso.
\newblock Su2: An open-source suite for multiphysics simulation and design.
\newblock {\em Aiaa Journal}, 54(3):828--846, 2016.

\bibitem{geuzaine2009gmsh}
Christophe Geuzaine and Jean-Fran{\c{c}}ois Remacle.
\newblock Gmsh: A 3-d finite element mesh generator with built-in pre-and post-processing facilities.
\newblock {\em International journal for numerical methods in engineering}, 79(11):1309--1331, 2009.

\bibitem{viana2016tutorial}
Felipe~AC Viana.
\newblock A tutorial on latin hypercube design of experiments.
\newblock {\em Quality and reliability engineering international}, 32(5):1975--1985, 2016.

\bibitem{giurgiutiu2015structural}
Victor Giurgiutiu.
\newblock Structural health monitoring of aerospace composites.
\newblock pages 449--507, 2015.

\bibitem{hexcel2016hexply}
Hexcel Corporation.
\newblock Hexply{\textregistered} 8552 product datasheet.
\newblock 2016.

\bibitem{mindlin1951influence}
R.D. Mindlin.
\newblock Influence of rotatory inertia and shear on flexural motions of isotropic, elastic plates.
\newblock {\em Journal of Applied Mechanics}, 18(1):31--38, 1951.

\bibitem{berri2019real}
Pier~Carlo Berri, Matteo Davide~Lorenzo Dalla~Vedova, and Laura Mainini.
\newblock Real-time fault detection and prognostics for aircraft actuation systems.
\newblock In {\em AIAA Scitech 2019 Forum}, page 2210, 2019.

\end{thebibliography}

\end{document}